\documentstyle[fullname,palatino]{article}

\title{A Corpus-Based Investigation of Definite Description Use}

\author{Massimo Poesio \\
        \texttt{poesio@cogsci.ed.ac.uk}
        \and
        Renata Vieira \\
        \texttt{renata@cogsci.ed.ac.uk}\\[3ex]
        Centre for Cognitive Science\\
        The University of Edinburgh \\
        2 Buccleuch Place \\
        Edinburgh EH8 9LW, UK 
	}
\date{}

\newcommand{\ANAPHORICSAME}{{\bf anaphoric (same head)}} 
\newcommand{\ASSOCIATIVE}{{\bf associative}}
\newcommand{\COREFERENT}{{\bf co-referent}}  
\newcommand{\LARGERUNFAMILIAR}{{\bf larger situation~/~unfamiliar}} 
\newcommand{\LEXITEM}[1]{\LINGEX{#1}}
\newcommand{\LINGEX}[1]{{\em #1}}

\newcommand{\NEWTERM}[1]{{\bf #1}}

\newcommand{\NP}{\SHORT{np}}
\newcommand{\OUT}[1]{}
\newcommand{\SECREF}[1]{\S\ref{#1}}
\newcommand{\SECTION}[2]{\section{#1}\label{#2}}
\newcommand{\SHORT}[1]{{\sc #1}}
\newcommand{\SREF}[1]{(\ref{#1})}
\newcommand{\SUBSECTION}[2]{\subsection{#1}\label{#2}}

\newcommand{\TPAR}[1]{\noindent{\em #1}\ }

\newcounter{EEXAMPLE}
\renewcommand{\theEEXAMPLE}{\alph{EEXAMPLE}}

\newenvironment{AEXAMPLE}{\begin{list}{}
    {\topsep      0pt
     \partopsep   0pt
     \itemsep     .0ex
     \leftmargin  30pt
     
     \usecounter{EEXAMPLE}
     }\vskip-\lastskip}{\end{list}}
\newcommand{\ENUMA}[1]{\refstepcounter{equation}
\label{#1}\item[(\theequation)] \begin{AEXAMPLE}}
\newcommand{\ENDENUMA}{\end{AEXAMPLE}}
\newcommand{\EITEM}{\stepcounter{EEXAMPLE}\item[\theEEXAMPLE.]}
\newenvironment{EXAMPLE}{\begin{list}{}
    {\topsep      4pt
     \itemsep     .0ex
     \labelwidth  20pt
     \leftmargin  40pt
     \labelsep	   5pt
     
     }}{\end{list}}

\newcommand{\EOLDA}[1]{
\item[(\ref{#1})] \begin{AEXAMPLE}}

\newcommand{\ENEW}[1]{\refstepcounter{equation}
\label{#1}\item[(\theequation)\hspace{10pt}]}

\begin{document}

%

%


\maketitle

\vspace{-10pt}
\begin{abstract}
We present the results of a study of definite descriptions use in written
texts aimed at assessing the feasibility of annotating corpora with information
about definite description interpretation. We ran two experiments, in which
subjects were asked to classify the uses of definite descriptions in a corpus
of 33 newspaper articles, containing a total of 1412 definite descriptions. We
measured the agreement among annotators about the classes assigned to definite
descriptions, as well as the agreement about the antecedent assigned to those
definites that the annotators classified as being related to an antecedent in
the text. The most interesting result of this study from a corpus annotation
perspective was the rather low agreement (K=0.63) that we obtained using
versions of Hawkins' and Prince's classification schemes; better results
(K=0.76) were obtained using the simplified scheme proposed by Fraurud that
includes only two classes, first-mention and subsequent-mention. The agreement
about antecedents was also not complete.  These findings raise questions
concerning the strategy of evaluating systems for definite description
interpretation by comparing their results with a standardized annotation. From
a linguistic point of view, the most interesting observations were the great
number of discourse-new definites in our corpus (in one of our experiments,
about 50\% of the definites in the collection were classified as discourse-new,
30\% as anaphoric, and 18\% as associative/bridging) and the presence of
definites which did not seem to require a complete disambiguation.
\end{abstract}


\vspace{2pt}
\footnoterule
\fnsymbol{footnote}This paper will appear in {\em Computational Linguistics}.

\newpage

%

\SECTION{Introduction}{intro_section}

The work presented in this paper was inspired by the growing realization in the
field of computational linguistics of the need for an experimental evaluation
of linguistic theories---semantic theories, in our case.  The evaluation we are
considering typically takes the form of experiments in which humans subjects
are asked to annotate texts from a corpus (or recordings of spoken
conversations) according to a certain classification scheme, and the agreement
among their annotations is measured (see, e.g., \cite{passonneau&litman:ACL93}
or the papers in \cite{CL:97:empirical_discourse}). These attempts at an
evaluation are, in part, motivated by the desire to put these theories on a
more `scientific' footing by ensuring that the semantic judgments on which they
are based reflect the intuitions of a large number of speakers;\footnote{E.g.,
recent work in linguistics shows that the agreement with a theory's predictions
may be a matter of how well the actual behavior distributes around the
predicted behavior, rather than an all-or-nothing affair
\cite{bard&sorace:96}.} but experimental evaluation is also seen as a necessary
precondition for the kind of system evaluation done, e.g., in the Message
Understanding initiative ({\sc muc}), where the performance of a system is
evaluated by comparing its output on a collection of texts with a standardized
annotation of those texts produced by humans
\cite{chinchor&sundheim:95}. Clearly, a {\sc muc}-style evaluation presupposes
an annotation scheme on which all participants agree.

Our own concern are semantic judgments concerning the interpretation of noun
phrases with the definite article \LEXITEM{the}, that we will call
\NEWTERM{definite descriptions}, following
\cite{russell:descriptions}.\footnote{We will not be concerned with other cases
of definite noun phrases such as pronouns, or possessive descriptions; hence
the term definite description rather than the more general term definite {\sc
np}.} These noun phrases are one the most common constructs in
English,\footnote{The word \LEXITEM{the} is by far the most common word in the
Brown corpus ~\cite{francis&kucera:82}, the {\sc lob} corpus
\cite{johansson&hofland:89}, and the {\sc trains} corpus
\cite{heeman&allen:trains-corpus-93}.} and have been extensively studied by
linguists, philosophers, psychologists, and computational linguists
\cite{russell:on-denoting,christopherson:39,strawson:on-referring,%
clark:bridging,grosz:thesis,cohen(phil):thesis,hawkins:78,%
sidner:thesis,webber:thesis,clark&marshall:81,prince:81,heim:82,appelt:85,%
loebner:definites,kadmon:thesis,carter:book,bosch&geurts:89,%
neale:descriptions,kornfeld:book,fraurud:90,barker:thesis,dale:book,%
cooper:STASS3,kamp&reyle:93,poesio:STASS3}.

Theories of definite descriptions such as
\cite{christopherson:39,hawkins:78,webber:thesis,prince:81,heim:82} identify
two subtasks involved in the interpretation of a definite description: deciding
whether the definite description is related to an antecedent in the
text\footnote{We concentrated on written texts in this study. See discussion
below.}--which in turn may involve recognizing fairly fine-grained
distinctions--and, if so, identifying this antecedent. Some of these theories
have been cast in the form of classification schemes
\cite{hawkins:78,prince:92}, and have been used for corpus analysis
\cite{prince:81,prince:92,fraurud:90};\footnote{Both Prince's and Fraurud's
studies are analyses of the use of the whole range of definite {\sc np}, not
just of definite descriptions.}  yet, we are aware of no attempt at verifying
whether non linguistically trained subjects are capable of recognizing the
proposed distinctions, which is a precondition for using these schemes for
the kind of large-scale text annotation exercises which are necessary to
evaluate a system's performance as done in {\sc muc}. 

In the past two or three years, this kind of verification has been attempted
for other aspects of semantic interpretation: e.g., by
\cite{passonneau&litman:ACL93} for segmentation and by
\cite{kowtko&isard&doherty:92,carletta-et-al:97} for dialogue act
annotation. Our intention was to do the same for definite descriptions.  We ran
two experiments to test how good are naive subjects at doing the form of
linguistic analysis presupposed by current schemes for classifying definite
descriptions. (Where by `how good' here we mean `how much do they agree among
themselves', as commonly assumed in work of this kind.) Our subjects were asked
to classify the definite descriptions found in a corpus of natural language
texts according to classification schemes that we developed starting from the
taxonomies proposed by Hawkins \shortcite{hawkins:78} and Prince
\shortcite{prince:81,prince:92}, but which took into account our intention of
letting `naive' speakers perform the classification. Our experiments were also
designed to assess the feasibility of a system to process definite descriptions
on unrestricted text and to collect data that could be used for this
implementation. For both of these reasons, the classification schemes that we
tried differ in several respects from those adopted in prior corpus-based
studies such as \cite{prince:81,fraurud:90}.  Our study is also different from
these previous ones in that measuring the agreement among annotators became an
issue \cite{carletta:kappa}.

We used for the experiments a set of randomly selected articles from the Wall
Street Journal contained in the {\sc acl/dci cd-rom}, rather than a corpus of
transcripts of spoken language corpora such as the {\sc hcrc} {\sc MapTask}
corpus \cite{anderson(anne)-et-al:91} or the {\sc trains} corpus
\cite{heeman&allen:trains-corpus-93}. The main reason for this choice was to
avoid dealing with deictical uses of definite descriptions and with phenomena
such as reference failure and repair. A second reason was that we intended to
use computer simulations of the classification task to supplement the results
of our experiments, and we needed a parsed corpus for this purpose; the
articles we chose were all part of the Penn Treebank
\cite{marcus-et-al:treebank}.

The organization of the paper is as follows. We review two existing
classification schemes in section \SECREF{literature_section}; we then discuss
our two classification experiments in sections \SECREF{first_exp_section} and
\SECREF{second_exp_section}, respectively.  

\SECTION{Towards a Classification Scheme: 
         Linguistic Theories of Definite Descriptions}{literature_section}

When looking for an annotation scheme for definite descriptions, one is faced
with a wide range of options. On the one end of the spectrum there are mostly
descriptive lists of definite description uses such as those in
\cite{christopherson:39,hawkins:78}, whose only goal is to assign a
classification to all uses of definite descriptions. On the other end there are
highly developed formal analyses such as
\cite{russell:on-denoting,heim:82,loebner:definites,kadmon:thesis,%
neale:descriptions,barker:thesis,kamp&reyle:93}, in which the compositional
contribution of definite descriptions to the meaning of an utterance, as well
as their truth-conditional properties, are spelled out in detail. These more
formal analyses are concerned with questions such as the quantificational or
non-quantificational status of definite descriptions and the proper treatment
of presuppositions, but tend to concentrate on a subset of the full range of
definite description use. Among the more developed semantic analyses, some
identify \NEWTERM{uniqueness} as the defining property of definite descriptions
\cite{russell:on-denoting,neale:descriptions}, whereas others take
\NEWTERM{familiarity} as the basis for the analysis
\cite{christopherson:39,hawkins:78,heim:82,prince:81,kamp&reyle:93}. We will
say more about some of these analyses below.

Our choice of a classification scheme was in part dictated by the intended use
of the annotation, in part by methodological considerations. A crucial property
of an annotation used to evaluate the performance of a system is that it ought
to identify the anaphoric connections between discourse entities; this makes
familiarity-based analyses more attractive. From a methodological point of
view, it was important to choose an annotation scheme that (i) would make the
classification task doable by non-linguistically trained subjects, and (ii) had
already been applied to the task of corpus analysis. We felt that we could ask
naive subjects to assign each definite description to one of a few classes and
to identify its antecedent when appropriate; we also wanted an annotation
scheme that would characterize the whole range of definite description use, so
that we would not need to worry about eliminating definite descriptions from
our texts because `unclassifiable'.

For these reasons we chose Hawkins' list of definite description uses
\cite{hawkins:78} and Prince's taxonomy \cite{prince:81,prince:92} as our
starting point, and we developed from there two slightly different annotation
schemes, which allowed us to see whether it was better to describe the classes
to our annotators in a surface-oriented or a semantic fashion, and to evaluate
the seriousness of the problems with these schemes identified in the literature
(see, e.g., \cite{fraurud:90}). We discuss Hawkins' and Prince's taxonomies
next.

\subsection{The Christopherson / Hawkins' List  of Definite Description Uses}

The wide range of uses of definite descriptions was already highlighted in
\cite{christopherson:39}. In the third chapter of his book, Hawkins
\shortcite{hawkins:78} further develops and extends Christopherson's list. He
identifies the following classes, or `uses,' of definite descriptions:

\subsection*{Anaphoric Use}

These are definite descriptions that \NEWTERM{co-specify}\footnote{There are
some complex terminological problems when discussing anaphoric
expressions. Following standard terminology, we will use the term
\NEWTERM{referent} to indicate the object in the world that is contributed to
the meaning of an utterance by a definite description--e.g., we will say that
Bill Clinton is the referent of a referential use of the definite description
\LINGEX{the president of the USA in 1997}.  We will then say, following
Sidner's terminology \cite{sidner:thesis}, that a definite description
\NEWTERM{co-specifies} with its antecedent in a text, when such antecedent
exists, if the definite description and its antecedent denote the same
object. This is probably the most precise way of referring to the relation
between an anaphoric expression and its antecedent; note that two discourse
entities can co-specify without referring to any object in the world--e.g., in
\LINGEX{The (current) king of France is bald. He has a double chin, as well.},
\LINGEX{he} co-specifies with \LINGEX{the (current) king of France}, but this
latter expression does not refer to anything. However, since we will mostly be
concerned with referential discourse entities, we will often use the term
\NEWTERM{co-refer} instead of \NEWTERM{co-specify}.  Apart from this, we have
tried to avoid more complex issues of reference insofar as possible
\cite{donnellan:72,kripke:77,barwise&perry:83,neale:descriptions,%
kornfeld:book}.} with a discourse entity already introduced in the
discourse. The definite description may use the same descriptive predicate as
its antecedent, or any other capable of indicating the same antecedent (e.g., a
synonym, a hyponym, etc.).

\begin{EXAMPLE}
\ENUMA{anaphoric-use}
\EITEM Fred was discussing {\em an interesting book} in his class. I
  went to discuss {\em the book} with him afterwards. 
\EITEM Bill was working at {\em a lathe} the other day. All of a sudden
  {\em the machine} stopped turning.
\EITEM Fred was wearing {\em trousers}. {\em The pants} had a big patch
  on them.
\EITEM Mary {\em travelled} to Paris. {\em The journey} lasted six hours.
\EITEM {\em A man and a woman} entered restaurant. {\em The couple} 
	was received by a waiter.
\ENDENUMA
\end{EXAMPLE}

\subsection*{Immediate Situation Uses}

The next two uses of definite descriptions identified by Hawkins are
occurrences used to refer to an object in the situation of utterance. The
referent may be visible, or its presence may be inferred. The \NEWTERM{visible
situation use} occurs when the object referred to is visible to both speaker
and hearer, as in the following examples:

\begin{EXAMPLE}
\ENUMA{visible-situation}
\EITEM Please, pass me {\em the salt}.
\EITEM Don't break {\em the vase}.
\ENDENUMA
\end{EXAMPLE}

Hawkins classifies as \NEWTERM{immediate situation uses} those definite
descriptions whose referent is a constituent of the immediate situation in
which the use of the definite description is located, without necessarily being
visible:
\begin{EXAMPLE}
\ENUMA{immediate-situation}
\EITEM Beware of {\em the dog}.
\EITEM Don't feed {\em the pony}.
\EITEM You can put your coat on {\em the clothes peg}.
\EITEM Mind {\em  the step}.
\ENDENUMA
\end{EXAMPLE}

\subsection*{Larger Situation Uses}

Hawkins lists then two uses of definite descriptions characteristic of
situations in which the speaker appeals to the hearer's knowledge of entities
which exist in the non-immediate or larger situation of utterance---knowledge
they share by being members of the same community, for instance.

A definite description may rely on \NEWTERM{specific knowledge about the larger
situation}: this is the case in which both the speaker and the hearer know
about the existence of the referent, as in the example below, in which it is
assumed that speaker and hearer are both inhabitants of Halifax, a town which
has a gibbet at the top of Gibbet Street:

\begin{EXAMPLE}
\ENEW{large-situation:specific}{\em The Gibbet} no longer stands.
\end{EXAMPLE}
Specific knowledge is not, however, a  necessary part of the meaning of
larger situation uses of definite descriptions. While some hearers may have
specific knowledge about the actual individuals referred to by a definite
description, others may not. General knowledge about the existence of certain
types of objects in certain types of situations is sufficient. Hawkins
classifies those definite descriptions which depend on this knowledge as
instances of \NEWTERM{general knowledge in the larger situation use}. An
example is the following utterance in the context of a wedding:

\begin{EXAMPLE}
\ENEW{large-situation:general}Have you seen {\em the bridesmaids}?
\end{EXAMPLE}
Such a first-mention of \LEXITEM{the bridesmaids} is possible on the
basis of the knowledge that weddings typically have bridesmaids. In the same
way, a first-mention of \LEXITEM{the bride}, \LEXITEM{the church 
service}, or
\LEXITEM{the best man} would be possible.

\subsection*{Associative Anaphoric Use}

Speaker and hearer may have (shared) knowledge of the relations between certain
objects (the \NEWTERM{triggers}) and their components or attributes (the
\NEWTERM{associates}): associative anaphoric uses of definite descriptions
exploit this knowledge. Whereas in larger situation uses the trigger is the
situation itself, in the associative anaphoric use the trigger is an {\sc np}
introduced in the discourse.

\begin{EXAMPLE}
\ENUMA{associative}
\EITEM The man drove past our house in {\em a car}. 
{\em The exhaust fumes} were terrible.
\EITEM I am reading {\em a book about Italian history}. 
       {\em The author} claims that Ludovico il Moro wasn't a bad ruler. 
       {\em The content} is generally interesting. 
\EITEM I went to {\em a wedding} last weekend. {\em The bride} was a friend
	of mine. She baked {\em  the cake } herself.
\ENDENUMA
\end{EXAMPLE}

\subsection*{Unfamiliar Uses}

Hawkins classifies as \NEWTERM{unfamiliar} those definite descriptions which
are not anaphoric, do not rely on information about the situation of utterance,
and are not associates of some trigger in the previous discourse.  Hawkins
groups these definite descriptions in classes according to their syntactic and
lexical properties, as follows.
\vspace{10pt}

\TPAR{NP complements} One form of unfamiliar definite descriptions is
characterized by the presence of a complement to the head noun.

\begin{EXAMPLE}
\ENUMA{np-complement}
\EITEM Bill is amazed by {\em the fact that there is so much life on Earth}.
\EITEM The philosophical aphasic came to {\em the conclusion that language
  did not exist}.
\EITEM Fleet Street has been buzzing with {\em the rumour that the Prime
  Minister is going to resign}.
\EITEM I remember {\em the time when I was a little girl}.
\ENDENUMA
\end{EXAMPLE}

\TPAR{Nominal modifiers} The distinguishing feature of these phrases, according
to Hawkins, is the presence of a nominal modifier which refers to the class to
which the head noun belongs.

\begin{EXAMPLE}
\ENUMA{nominal-modifiers}
\EITEM I don't like {\em the colour red}.
\EITEM {\em The number seven} is my lucky number.
\ENDENUMA
\end{EXAMPLE}

\TPAR{Referent Establishing Relative Clauses} Relative clauses may establish a
referent for the hearer without a previous mention, when the relative clause
refers to something mutually known.

\begin{EXAMPLE}
\ENUMA{referent-establishing}
\EITEM What's wrong with {\bf Bill}? 
       Oh, {\em the woman} {\bf  he} {\em went out with last
       night} was nasty to him. (But: ?? Oh, {\em the woman} was nasty to 
him.)
\EITEM {\em The box (that is)} {\bf over there}
\ENDENUMA
\end{EXAMPLE}

\TPAR{Associative clauses} Some definite descriptions can be seen as cases of
bridging references in which both the trigger and the associate are
specified. The modifiers of the head noun specify the set of objects with which
the referent of the definite description is associated. 

\begin{EXAMPLE}
\ENUMA{associative-clauses}
\EITEM I remember {\em the beginning of the war} very well.
\EITEM There was a funny story on {\em  the front page of the Guardian} this
  morning.
\EITEM ... {\em the bottom of the sea}.
\EITEM ... {\em the fight during the war}.
\ENDENUMA
\end{EXAMPLE}

\subsection*{Unexplanatory Modifiers Use} 

Finally, Hawkins lists a small number of modifiers which require the use of
\LEXITEM{the}: 

\begin{EXAMPLE}
\ENUMA{unexplanatory}
\EITEM My wife and I share {\em the same secrets}.
\EITEM {\em The first person to sail to America} was an Icelander.
\EITEM {\em The fastest person to sail to America} ...
\ENDENUMA
\end{EXAMPLE}

\subsection{The Semantics of Definite Descriptions}

Some of the classes in the Christopherson / Hawkins classification are
specified in a semantic fashion; other classes are defined in purely syntactic
terms. It is natural to ask what these uses of definite descriptions have in
common from a semantic point of view: for example, is there a connection
between the `unfamiliar' and the `unexplanatory' uses of definite descriptions
and the other uses?  (The unfamiliar uses with associative clauses seem related
to the associative anaphoric ones, and both to the uses based on referent
establishing relative clauses.) Many authors, including Hawkins himself, have
attempted to go beyond the purely descriptive list just discussed.

One group of authors have identified \NEWTERM{uniqueness} as the defining
property of definite descriptions. This idea goes back to
\cite{russell:on-denoting}, and is motivated by larger situation definite
descriptions such as \LINGEX{the pope} and by some cases of unexplanatory
modifier use such as \LINGEX{the first person to sail to America}. The
hypothesis was developed in recent years
\cite{kadmon:thesis,neale:descriptions,cooper:STASS3}, in particular to address
the problem of `uniqueness within small situations'.\footnote{L{\"{o}}bner
generalizes this idea to good results in \cite{loebner:definites}; we will
return on this work later.}

Another line of research is based on the observation that many of the uses of
definite descriptions listed by Hawkins have one property in common: the
speaker/writer is making some assumptions about what the hearer already
knows. Speaking very loosely, we might say that the speaker assumes that the
hearer is able to `identify' the referent of the definite description. This is
also true of some of the uses Hawkins classified as `unfamiliar': for example,
of his `nominal modifiers' and `associative clause' classes. Attempts at making
this intuition more precise include Christopherson's familiarity theory
\shortcite{christopherson:39}, Strawson's presuppositional theory of definite
descriptions \cite{strawson:on-referring}, Hawkins' own location theory
\cite{hawkins:78} and its revision, Clark and Marshall's theory of definite
reference and mutual knowledge \cite{clark&marshall:81}, as well as more formal
proposals such as \cite{heim:82}.

Neither the uniqueness nor the familiarity approach have yet succeeded in
providing a satisfactory account of all uses of definite descriptions
\cite{fraurud:90,birner&ward:94}. However, the theories based on familiarity
address more directly the main concern of {\sc nlp} system designers, which is
to identify the connections between discourse entities. Furthermore, the prior
corpus-based studies of definite descriptions use that we are aware of
\cite{prince:81,fraurud:90,prince:92} are based on theories of this type.  For
both of these reasons, we adopted semantic notions introduced in
familiarity-style accounts in designing our experiments---in particular,
distinctions introduced in Prince's taxonomy.

\subsection{Prince's Classification of Noun Phrases}

\begin{sloppypar}
Prince studied in detail the connection between a speaker / writer's
assumptions about the hearer or reader and the linguistic realization of noun
phrases \cite{prince:81,prince:92}.  She criticizes as too simplistic the
binary distinction between `given' and `new' discourse entities that is at the
basis of most previous work on familiarity, and proposes a much more detailed
taxonomy of `givenness'---or, as she calls it, \NEWTERM{assumed
familiarity}---meant to address this problem. Also, Prince's analysis of noun
phrases is closer than the Christopherson / Hawkins' taxonomy to a
classification of definite descriptions on purely semantic terms: e.g., she
relates `unfamiliar' definites based on referent-establishing relative clauses
with Hawkins' associative clause and associative anaphoric uses.\footnote{Clark
and Marshall \shortcite{clark&marshall:81} also proposed a revision of Hawkins'
theory that merges some of the classes on semantic grounds.}
\end{sloppypar}

\subsection*{Hearer New / Hearer Old}

One factor affecting the choice of a noun phrase, according to Prince, is
whether a discourse entity is old or new with respect to the hearer's
knowledge. A speaker will use a proper name or a definite description when he
or she assumes that the addressee already knows the entity whom the speaker is
referring to, as in \SREF{prince:HO} and \SREF{prince:HO:def}.

\begin{EXAMPLE}
\ENEW{prince:HO} I'm waiting for it to be noon so I can call \LINGEX{Sandy
Thompson}. 
\ENEW{prince:HO:def} Nine hundred people attended \LINGEX{the Institute}.
\end{EXAMPLE}
On the other hand, if the speaker believes that  the addressee does not know
of Sandy Thompson, an indefinite will be used:

\begin{EXAMPLE}
\ENEW{prince:HN}I'm waiting for it to be noon so I can call \LINGEX{someone in
California}.  
\end{EXAMPLE}

\subsection*{Discourse New / Discourse Old}

In addition, discourse entities can also be new or old with respect to the
discourse model: an {\NP} may refer to an entity that has already been `evoked'
in the current discourse, or it may evoke an entity which has not been
previously mentioned. `Discourse novelty' is distinct from `Hearer novelty':
both Sandy Thompson in \SREF{prince:HO} and the \LINGEX{someone in California}
mentioned in \SREF{prince:HN} may well be discourse-new even if only the second
one will be hearer-new. On the other hand, for an entity being discourse old
entails it being hearer old. In other words, in Prince's theory the notion of
`familiarity' is split in two: familiarity with respect to the discourse, and
familiarity with respect to the hearer. Either type of familiarity can license
the use of definites: Hawkins' anaphoric uses of definite descriptions are
cases of noun phrases referring to discourse-old discourse entities, whereas
his `larger situation' and `immediate situation' uses are cases of noun phrases
referring to discourse-new, hearer-old entities.\footnote{In Clark and
Marshall's \shortcite{clark&marshall:81} terminology, one would say that
different co-presence heuristics can be used to establish mutual knowledge.}

\subsection*{Inferrables}

The uses of definite descriptions that Hawkins called associative anaphoric,
such as \LINGEX{a book}\ldots \LINGEX{the author}, are not discourse-old or
even hearer-old, but they are not entirely new, either; as Hawkins pointed out,
the hearer is assumed to be capable to infer their existence. Prince called
these discourse entities \NEWTERM{inferrables}. (This is the class of definite
descriptions for which Clark \shortcite{clark:bridging} used the term
\NEWTERM{bridging references}.)

\subsection*{Containing Inferrables}

Finally, Prince proposes a category for noun phrases that are like inferrables,
but whose connection with previous hearer's knowledge is specified as part of
the noun phrase itself---her example is \LINGEX{the door of the Bastille} in
the following example:

\begin{EXAMPLE}
\ENEW{prince:CI} The door of the Bastille was painted purple. 
\end{EXAMPLE}
At least three of the `unfamiliar uses' of Hawkins---NP complements,
referent-establishing relative clauses, and associative clauses---fall in this
category. (See also \cite{clark&marshall:81}.)

\subsection{Some Remarks about Coverage}

Perhaps the most important question concerning a classification scheme is its
coverage. The two taxonomies we have just seen are largely satisfactory in this
respect, but a couple of issues are worth mentioning.

First of all, Prince's taxonomy does not give us a complete account of the
licensing conditions for definite descriptions. Of the uses mentioned by
Hawkins, the unfamiliar definites with unexplanatory modifiers and {\sc np}
complements need not satisfy any of the conditions that license the use of
definites according to Prince: these definites are not necessarily
discourse-old, hearer-old, inferrables, or containing inferrables. These uses
fall outside of Clark and Marshall's classification, as well.

Secondly, none of the classification schemes just discussed, nor any of the
alternatives proposed in the literature, consider so-called {\em generic} uses
of definite descriptions, such as the use of \LINGEX{the tiger} in the generic
sentence \LINGEX{The tiger is a fierce animal that lives in the jungle}. The
problem with these uses is that the very question of whether the `referent' is
familiar or not seems misplaced---these uses are not `referential'. A problem
related to the one just mentioned is that certain uses of definite descriptions
are ambiguous between a {\em referential} and an {\em attributive}
interpretation \cite{donnellan:72}. The sentence \LINGEX{The first person to
sail to America was an Icelander}, for example, can have two interpretations:
the writer may either refer to a specific person, whose identity may be
mutually known to both writer and reader; or he/she may be simply expressing a
property that is true of the first person to sail to America, whoever that
person happened to be. This ambiguity does not seem to be possible with all
uses of definite descriptions: e.g., \LINGEX{pass me the salt} only seem to
have a referential use. Again, the schemes we have presented do not consider
this issue. The question of how to annotate generic uses of definite
descriptions or uses that are ambiguous between a referential and an
attributive use will not be addressed in this paper.

\subsection{Fraurud's Study}

A second problem with the classification schemes we have discussed was raised
by Fraurud in her study of definite {\NP}s in a corpus of Swedish text
\cite{fraurud:90}. Fraurud introduced a drastically simplified classification
scheme based on two classes only: \NEWTERM{subsequent mention}, corresponding
to Hawkins' anaphoric definite descriptions and Prince's discourse-old, and
\NEWTERM{first-mention}, including all other definite descriptions.

Fraurud simplified matters in this way because she was primarily interested in
verifying the empirical basis for the claim that familiarity is the defining
property of definite descriptions; she also observed, however, that some of the
distinctions introduced by Hawkins and Prince led to ambiguities of
classification. For example, she observed that the reader of a Swedish
newspaper can equally well interpret the definite description \LINGEX{the king}
in an article about Sweden by reference to the larger situation or to the
content of the article.

We took into account Fraurud's observations in designing our experiments, and
we will compare our results to hers below.

\SECTION{A First Experiment in Classification}{first_exp_section}

For our first experiment at evaluating subjects' performance at the
classification task, we developed a taxonomy of definite description uses based
on the schemes discussed in the previous section, preliminarily tested the
taxonomy by annotating the corpus ourselves, and then asked two annotators to
do the same task. This first experiment is described in the rest of this
section. We explain, first, the classification we developed for this
experiment, then the experimental conditions, and finally discuss the results.

\SUBSECTION{The First Classification Scheme}{first_classification_section}

The annotation schemes for noun phrases proposed in the literature fall in one
of two categories. On the one hand, we have what we might call `labeling'
schemes, most typically used by corpus linguists, which involve assigning to
each noun phrase a class such as those discussed in the previous section; the
schemes used by Fraurud and Prince fall in this category.  On the other hand,
there are what we might call `linking' schemes, concerned with identifying the
links between the discourse entity or entities introduced by a noun phrase and
other entities in the discourse; the scheme used in {\sc muc}-6 is of this
type.

In our experiments, we tried both a purely labeling scheme and a mixture of a
labeling and a linking scheme. We also tried two slightly different taxonomies
of definite descriptions, and we varied the way membership in a class was
defined to the subjects. Both taxonomies were based on the schemes proposed by
Hawkins and Prince, but we introduced some changes in order, first, to find a
scheme that would be easily understood by individuals without previous
linguistic training and would lead to maximum agreement among the classifiers;
and second, to make the classification more useful for our goal of feeding the
results into an implementation.

\begin{sloppypar}
In the first experiment, we used a labeling scheme, and the classes 
were introduced to the subjects with reference to the surface characteristics
of the definite descriptions. (See below and Appendix A.) The taxonomy we used
in this experiment is a simplification of Hawkins' scheme, to which we made
three main changes. First of all, we separated those anaphoric descriptions
whose antecedents have the same descriptive content as their antecedent (which
we will call \ANAPHORICSAME) from other cases of anaphoric descriptions in
which the association is based on more complex forms of lexical or commonsense
knowledge (synonyms, hypernyms, information about events, etc.). We grouped
these latter definite descriptions with Hawkins' associative descriptions in a
class that we called \ASSOCIATIVE. This was done in order to see how much need
there is for complex lexical inferences in resolving anaphoric definite
descriptions, as opposed to simple head matching.
\end{sloppypar}

Secondly, we grouped together all the definite descriptions which introduce a
novel discourse entity not associated to some previously established object in
the text, i.e., that were discourse-new in Prince's sense. This class, that we
will call \LARGERUNFAMILIAR, includes both definite descriptions that exploit
situational information (Hawkins' \NEWTERM{larger situation} uses) and
discourse-new definite descriptions introduced together with their links or
referents (\NEWTERM{unfamiliar}).  This was done becase of Fraurud's
observation that distinguishing the two classes is generally difficult
\cite{fraurud:90}. Third, we did not include a class for immediate situation
uses, since we assumed they would be rare in written text.\footnote{This was
indeed the case, but we did observe a few instances of an interesting kind of
immediate situation use. In these cases, the text is describing the immediate
situation in which the writer is, and the writer apparently expects the reader
to reconstruct this situation:

\begin{EXAMPLE}
\ENEW{immediate:table}
 ``And you didn't want me to buy earthquake insurance'',  says
  Mrs. Hammack, reaching across {\em the table} and gently tapping 
his hand.
\ENEW{immediate:head} 
 ``I will sit down and talk some of the problems out, but take on the
 political system ?  Uh-uh'',  he says with a shake of {\em the head}.
\end{EXAMPLE}
}
We also introduced a separate class of \NEWTERM{idioms} including indirect
references, idiomatic expressions and metaphorical uses, and we allowed our
subjects to mark definite descriptions as \NEWTERM{doubts}.  

To summarize, the classes used in this experiment were as follows.

\subsection*{I. Anaphoric same head} 

This class includes uses of definite descriptions which refer back to an
antecedent introduced in discourse; it differs from Hawkins' `anaphoric use' or
Prince's `textually evoked' classes because it only includes
definite-antecedent pairs with the same head noun.

\begin{EXAMPLE} 
\ENEW{anaphora_sh}
 Grace Energy just two weeks ago hauled {\em a rig} here 500 miles from
Caspar, Wyo., to drill the Bilbrey well, a 15,000-foot, \$
1-million-plus natural gas well.
 {\em The rig} was built around 1980, but has drilled only two wells,
the last in 1982.
\end{EXAMPLE}

\subsection*{II. Associative} 

We assigned to this class those definite descriptions that stand in an
anaphoric or associative anaphoric relation with an antecedent explicitly
mentioned in the text, but that are not identified by the same head noun as
their antecedent. This class includes Hawkins' associative anaphoric definite
descriptions and Prince's inferrables, as well as some definite descriptions
that would be classified as anaphoric by Hawkins and as textually evoked in
\cite{prince:81}.)  Recognizing the antecedent of these definite descriptions
involves at least knowledge of lexical associations, and possibly general
commonsense knowledge.\footnote{See
\cite{loebner:definites,barker:thesis,poesio:SALT4} for discussions of lexical
conditions on bridging references.}

\begin{EXAMPLE} 
\ENUMA{associative-ext}

\EITEM 
With all this, even the most wary oil men agree {\em something has 
changed}.
``It doesn't appear to be getting worse''.
``That in itself has got to cause people to feel a little more 
optimistic,'' says Glenn Cox, the president of Phillips Petroleum Co.
Though modest, {\em the change} reaches beyond the oil patch, too.

\EITEM 
Toni Johnson pulls a tape measure across the front of what was once
{\em a stately Victorian home}.
A deep trench now runs along its north wall, exposed when {\em  the 
house} lurched two feet off its foundation during last week's 
earthquake.

\EITEM 
Once inside, she spends nearly four hours measuring and
diagramming each room in {\em the 80-year-old house}, gathering enough 
information to estimate what it would cost to rebuild it.
 While she works inside, a tenant returns with several friends to collect 
furniture and clothing.
 One of the friends sweeps broken dishes and shattered glass
from a countertop and starts to pack what can be salvaged from {\em the 
kitchen}.

\ENDENUMA
\end{EXAMPLE}

\subsection*{III. Larger situation/unfamiliar}  

This class includes Hawkins' larger situation uses of definite descriptions
based on specific and general knowledge (discourse-new, hearer-old in Prince's
terms) as well as his unfamiliar uses (many of which correspond to Prince's
containing inferrables).

\begin{EXAMPLE} 
\ENUMA{largesit-ext}

\EITEM 
Out here on {\em the Querecho Plains of New Mexico}, however, the 
mood is
more upbeat trucks rumble along the dusty roads and burly men in hard
hats sweat and swear through the afternoon sun.

\EITEM 
Norton Co. said net income for {\em the third quarter}  fell 6 \% to
\$ 20.6 million, or 98 cents a share, from \$ 22 million, or \$ 1.03 a
share.

\EITEM 
For the Parks and millions of other young Koreans, the long-cherished 
dream of home ownership has become a cruel illusion.
For {\em the government}, it has become a highly volatile political 
issue.

\EITEM  
About the same time, {\em the Iran-Iraq war}, which was roiling oil 
markets, ended.

\ENDENUMA
\end{EXAMPLE}

\subsection*{ IV. Idiom}  

This class includes indirect references, idiomatic expressions and metaphorical
uses.

\begin{EXAMPLE} 
\ENEW{idiom} 

A recession or new OPEC blowup could put oil markets right back in {\em 
the soup}.
\end{EXAMPLE}

\subsection{Experimental Conditions}

First of all, we classified ourselves the definite descriptions included in 20
randomly chosen articles from the Wall Street Journal contained in the subset
of the Penn Treebank corpus included in the ACL/DCI CD-ROM.\footnote{The texts
in question are w0203, w0207, w0209, w0301, w0305, w0725, w0760, w0761, w0765,
w0766, w0767, w0800, w0803, w0804 w0808, w0820, w1108, w1122, w1124, and
w1137.}  All together, these articles contain 1040 instances of definite
description use. The results of our analysis are summarized in Table
\ref{table-1}.

\begin{table}[h]
\begin{tabular}{|l|l|l|}
{\bf Class} & Total Number & Percentage of the total \\
\hline\hline
{\bf I.~Anaphoric s. h.}    &  304 & 29.23\% \\
\hline
{\bf II.~Associative}       &  193 & 18.55\% \\
\hline
{\bf III.~LS/Unfamiliar}    &  503 & 48.37\% \\
\hline
{\bf IV.~Idiom}             &  26  &  2.50\% \\
\hline
{\bf V.~Doubt}              &  14  &  1.35\% \\
\hline
{\bf Total}                 & 1040  &  100  \\
\hline
\hline
\end{tabular}
\caption{\label{table-1}  Classification by the authors of the definite 
descriptions in the first corpus}
\end{table}
Next, we asked 2 subjects to perform the
same task. Our two subjects in this first experiment were graduate students in
Linguistics.  The two subjects were given the instructions in Appendix A. They
had to assign each definite description to one of the classes described in
\SECREF{first_classification_section}: I. {\ANAPHORICSAME}, II.
{\ASSOCIATIVE}, III. {\LARGERUNFAMILIAR}, and IV. {\bf idiom}. The subjects
could also express V. `doubt' about the classification of the definite
description. Since the classes I-III are not mutually exclusive, we instructed
the subjects to resolve conflicts according to a preference ranking, i.e., to
choose a class with higher preference when two classes seemed equally
applicable. The ranking was (from most preferred to least preferred): 1)
{\ANAPHORICSAME}, 2) {\LARGERUNFAMILIAR}, and 3) {\ASSOCIATIVE}. The annotators
were given one text to familiarize themselves with the task before starting
with the annotation proper.

\begin{table}[h]
\begin{tabular}{|l|l|l|}
{\bf Class}                     & Total Number & Percentage of the total \\
\hline\hline
{\bf I.Anaphoric s. h.}                   &  294  & 28.27\% \\
\hline
{\bf II.Associative}                     &  160  & 15.38\% \\
\hline
{\bf III.Unfamiliar/Larger Situation}      &  546  & 52\% \\
\hline
{\bf IV.Idiom}                            &  39   &  3.75\% \\
\hline
{\bf V.Doubt}                             &  1    &  0.09\% \\
\hline
{\bf Total}                               &  1040 &  100\%  \\
\hline
\hline
\end{tabular}
\caption{\label{ann-A-table} Classification of definite   descriptions
according to Annotator A.}
\end{table}

\subsection{Results}

\subsection*{The distribution of definite descriptions in classes}

The results of the first annotator (henceforth, `Annotator A') are shown in
Table \ref{ann-A-table}, and those of the second annotator (henceforth,
`Annotator B') in Table \ref{ann-B-table}.

\begin{table}
\begin{tabular}{|l|l|l|}
{\bf Class}                     & Total Number & Percentage of the total \\
\hline\hline
{\bf I.Anaphoric s. h.}                  &  332  & 31.92\% \\
\hline
{\bf II.Associative}                    &  150  & 14.42\% \\
\hline
{\bf III.Unfamiliar/Larger Situation}     &  549  & 52.78\% \\
\hline
{\bf IV.Idiom}                           &  2    &  0.19\% \\
\hline
{\bf V.Doubt}                            &  7    &  0.67\% \\
\hline
{\bf Total}                              &  1040 &  100\%  \\
\hline
\hline
\end{tabular}
\caption{\label{ann-B-table} Classification of definite   descriptions
according to Annotator B.}
\end{table}
As the tables indicate, the annotators and us assigned approximately the
same percentage of definite descriptions to each of the five classes; however,
the classes do not always include the same elements. This can be gathered by
the confusion matrix in Table \ref{cm-table}, where an entry $m_{x,y}$
indicates the number of definite descriptions assigned to class $x$ by subject
A and to class $y$ by subject B.

\begin{table}
\begin{tabular}{|l|l|l|l|l|l|l|}
~~~~{\bf B}&{\bf I.}& {\bf II.}&{\bf III.}&{\bf IV.}& {\bf V.}&{\bf Total B}\\
{\bf A}    &       &      &      &     &     &    \\
\hline
\hline
{\bf I. Anaphoric}& {\bf 274} &  26   &   32    &  0   &  0   &   332 \\      
\hline 
{\bf II. Associative}   &   9   &  {\bf 97}&    44   &  0   &  0   &   150 \\
\hline
{\bf III. LS/Unfamiliar}&   8   &   37   & {\bf 465}&  38  &  1   &549 \\     
\hline 
{\bf IV. Idiom}         &   0   &   0    &    1    &  {\bf 1}&  0   &  2 \\    
\hline
{\bf V. Doubt}          &   3   &   0    &    4    &   0  & {\bf 0}   &  7 \\
\hline    
{\bf Total A} &  294   & 160   &   546   &  39  &  1   &  1040 \\
\hline
\hline
\end{tabular}
\caption{\label{cm-table} Confusion matrix of A and B's classifications.}
\end{table}

In order to measure the agreement in a more precise way, we used the so-called
\NEWTERM{Kappa Statistic} \cite{siegel&castellan:88}, recently proposed by
Carletta as a measure of agreement for discourse analysis
\cite{carletta:kappa}. We also used a measure of per-class agreement that we
introduced ourselves. We discuss these results below, after reviewing briefly
how K is computed.

\subsection*{The Kappa Statistic}

Kappa is a test suitable for the cases when the subjects have to assign items
to one of a set of non-ordered classes. The test computes a coefficient `K' of
agreement among coders which takes into account the possibility of chance
agreement. It is dependent on the number of coders, number of items being
classified, and number of choices of classes to be ascribed to items. 

The kappa coefficient of agreement between $k$ annotators is defined as

\begin{EXAMPLE}
\ENEW{kappa} $K = \frac{P(A) - P(E)}{1 - P(E)}$
\end{EXAMPLE}
where $P(A)$ is the proportion of times the annotators agree and $P(E)$ is the
proportion of times that we would expect the annotators to agree by chance.
When there is complete agreement among the raters, K = 1; if there is no
agreement other than that expected by chance, K = 0. According to Carletta, in
the field of content analysis---where the Kappa statistic originated---$K> 0.8$
is generally taken to indicate good reliability, whereas $0.68 \leq K < 0.8$
allows tentative conclusions to be drawn.

We will illustrate the method for computing K proposed in
\cite{siegel&castellan:88} by means of an example from one of our texts, shown
in Table \ref{table-2}.

\begin{center}
\begin{table}[h]
{\small 
\begin{tabular}{|l|l|l|l|l|}
\hline
& & & &  \\
{\bf Definite description}&{\bf ASH}& {\bf ASS}& {\bf LSU}& {\bf S}\\
& & & &  \\
\hline
1. the third quarter 
  &  0 &  0 &  3   &    1\\
\hline
2. the abrasives, engineering materials & & & &\\
and petroleum services concern
    & 0  & 2  & 1    & 0.33\\
\hline
3.  The company 
   & 0  & 3  & 0  & 1\\
\hline
4. the year-earlier quarter
   & 0  & 2  & 1 &  0.33\\
\hline
5. the tax credit 
  &  3 &  0  &   0  & 1\\
\hline
6. the engineering materials segment 
   & 1  & 1  & 1    & 0\\
\hline
7. the possible sale of all or part of & & & &\\
Eastman Christensen 
   &  0 &  0 &  3 &  1\\
\hline
8. the nine months 
   &  0  &  0  & 3 &  1\\
\hline
9.  the year-earlier period 
    & 0 &  2 &  1 &  0.33\\
\hline
10. the company 
    &  3 &  0 &  0 &  1\\
\hline
11. the company
   & 3 &  0 &  0  &  1\\
\hline
12.  the company 
   & 3 &    0 &  0 &  1\\
\hline
13.   the company 
   & 3 &  0 &    0  & 1\\
\hline

\hline 
& & & & \\
{\bf N}=13 & {\bf ASH}=16 & {\bf ASS}=10   & {\bf LSU}=13  & {\bf Z}=10\\

\hline 
\end{tabular}}
\caption{\label{table-2}  Exemplification of the Kappa test}
\end{table}
\end{center}

The first column in Table \ref{table-2} ({\bf Definite description}) shows the
definite description being classified.  The columns {\bf ASH}, {\bf ASS}, and
{\bf LSU} stand for the classification options presented to the subjects
({\ANAPHORICSAME, \ASSOCIATIVE, and \LARGERUNFAMILIAR, respectively).  The
numbers in each $n_{ij}$ entry of the matrix indicate the number of classifiers
that assigned the description in row $i$ to the class in column $j$. The final
column (labelled {\bf S}) represents the percentage agreement for each definite
description; we explain below how this percentage agreement is calculated.  The
last row in the table shows the total number of descriptions ({\bf N}), the
total number of descriptions assigned to each class and, finally, the total
percentage agreement for all descriptions ({\bf Z}).

The equations for computing S$_i$, PE, PA, and K are shown in Table
\ref{formulas_table}. In these formulas, $c$ is the number of coders; $S_i$ the
percentage agreement for description $i$ (we show S$_1$ and S$_2$ as examples);
$m$ the number of categories; $T$ the total number of classification judgments;
$PE$ the percentage agreement expected by chance; $PA$ the total agreement, and
$K$ is the Kappa coefficient.

\begin{table}[h]
\noindent \rule{\textwidth}{0.1ex}

\begin{tabbing}
$S_{i}$ \=$= 1 / c(c-1) * \sum_{j=1}^{m} n_{ij}(n_{ij}-1)$\\[1ex]
$S_{1}$ \>$= 1 / 3(2) * [0 + 0 + 3(2) ] = (1/6) * 6 = 1$\\
$S_{2}$ \>$= 1/6 * [0 + 2(1) + 1(0) ] = (1/6) * 2 = 0.33$\\[4ex]

$T=39$\\[4ex]

$PE$ \=$= (ASH/T)^{2} + (ASS/T)^{2} + (LSU/T)^{2} $\\[1ex]
     \>$= (16/39)^{2} + (10/39)^{2} + (13/39)^{2} $\\
     \>$= 0.17 + 0.07 + 0.11 = 0.35$\\[4ex]

$PA = Z/N = 10/13 = 0.77$\\[4ex]

$K = (PA - PE)/(1-PE) = (0.77 - 0.35) / (1 - 0.35) = 0.42/0.65  = 0.65$
\end{tabbing}
\noindent \rule{\textwidth}{0.1ex}
\caption{\label{formulas_table} Computing the K coefficient of agreement.}
\end{table}

\subsection*{Value of K for the first experiment}

For the first experiment, K=0.68 if we count idioms  as a class,
K=0.73 if we take them out. The overall coefficient of agreement between the
two annotators and our own analysis is K=0.68 if we count idioms, K=0.72 if we
ignore them.

\subsection*{Per-class agreement}

K gives a `global' measure of agreement. We also wanted to measure the
agreement per class, i.e., to understand where annotators agreed the most and
where they disagreed the most. The confusion matrix does this to some extent,
but only works for two annotators---and therefore, for example, we couldn't use
it to measure agreement on classes between the two annotators and ourselves.

We computed what we called `per-class percentage of agreement' for three coders
(the two annotators and ourselves) by taking the proportion of pairwise
agreements relative to the number of pairwise comparisons, as follows: whenever
all three coders ascribe a description to the same class, we count 6 pairwise
agreements out of 6 pairwise comparisons for that class - 100\%. If two coders
ascribe a description to class 1 and the other coder to class 2, we count two
agreements in four comparisons for class 1 (50\%) and no agreement for class 2
(0\%). The rates of agreement for each class thus obtained are presented in
Table \ref{table-each-class-1}. The figures indicate better agreement on
anaphoric same-head and larger situation / unfamiliar definite descriptions,
worse agreement on the other classes. (In fact, the percentages for idioms and
doubts are very low; but these classes are also too small to allow us to draw
any conclusions.)

\begin{table}
{\small 
\begin{tabular}{|l|l|l|l|l|l|}
{\bf Class}            & Total& Comparisons& Agree & Disagree&\% Agreement\\
\hline
{\bf I. Anaphoric s. h.}&  930 & 1860 & 1646 & 214 & 88\% \\
\hline
{\bf II. Associative}    &  503 & 1006 & 596  & 410 & 59\% \\
\hline
{\bf III. LS/Unfamiliar}  & 1598 & 3196 & 2684 & 512 & 84\% \\
\hline
{\bf IV. Idiom}          &   67 & 134  & 42   & 92  & 31\%\\
\hline
{\bf V. Doubt}           & 22   & 44   & 2    & 42  & 4\% \\
\hline
\end{tabular}
}
\caption{\label{table-each-class-1} Per-class agreement in Experiment 1.}
\end{table}

\subsection{Discussion of the results}

\subsection*{Distribution} 

One of the most interesting results of this first experiment is that a large
proportion of the definite descriptions in our corpus (48.37\%, according to
our own annotation; more, according to our two annotators) are not related to
an antecedent previously introduced in the text. Surprising as it may seem,
this finding is in fact just a confirmation of the results of other
researchers. \cite{fraurud:90} reports that 60.9\% of definite descriptions in
her corpus of 11 Swedish texts are `first-mention', i.e., do not co-refer with
an entity already evoked in the text;\footnote{As mentioned above, Fraurud's
first-mention class consists of Prince's discourse-new, inferrables, and
containing inferrables.}  \cite{gallaway:96} found a distribution similar to
ours in (English) spoken child language.


\subsection*{Disagreements Among Annotators} 

The second notable result was the relatively low agreement among annotators.
The reason for this disagreement was not so much annotators' errors as the
fact, already mentioned, that the classes are not mutually exclusive. The
confusion matrix in Table \ref{cm-table} indicates that the major classes of
disagreements were definite descriptions classified by annotator A as larger
situation and by annotator B as associative, and viceversa. One such example is
\LINGEX{the government} in \SREF{korea:ex}. This definite description could be
classified as larger situation because it refers to the government of Korea,
and presumably the fact that Korea has a government is shared knowledge; but it
could also be classified as being associative on the predicate
\LINGEX{Koreans}.\footnote{As discussed above, this problem with Hawkins' and
Prince's classification schemes had already been noted by Fraurud---e.g.,
\cite{fraurud:90}, page 416.}

\begin{EXAMPLE}
\ENEW{korea:ex}
For the Parks and millions of other young Koreans, the long-cherished dream of
home ownership has become a cruel illusion.  For \LINGEX{the government}, it
has become a highly volatile political issue.
\end{EXAMPLE}
We will analyze the reasons for the disagreement in more detail in relation to
our second experiment, in which we also asked the annotators to indicate
the antecedent of definite descriptions (see below).

\subsection*{Surface Indicators of Discourse Novelty}

Examining the annotations produced in this experiment, we were able to confirm
the correlation observed by Hawkins between the syntactic structure of certain
definite descriptions and their classification as discourse-new. Factors that
strongly suggest that a definite description is discourse-new (and in fact,
presumably hearer-new as well) include the presence of modifiers such as
\LINGEX{first} or \LINGEX{best}, and of a complement for {\NP}s of the form
\LINGEX{the fact that \ldots} or \LINGEX{the conclusion that
\ldots}.\footnote{We will discuss an explanation for this correlation suggested
in \cite{loebner:definites}.}  Post-nominal modification of any type is also a
strong indicator of discourse novelty, suggesting that most post-nominal
clauses serve to establish a referent in the sense discussed in the previous
section.  In addition, we observed a previously unreported (to our knowledge)
correlation between discourse-novelty and syntactic constructions such as
appositions, copular constructions, and comparatives. The following examples
from our corpus illustrate the correlations just mentioned:

\begin{EXAMPLE} 
\ENUMA{unfamiliar-ext}

\EITEM  
Mr. Ramirez, who arrived late at the Sharpshooter with his crew
because he had started early in the morning setting up tanks at
another site, just got {\em the first raise he can remember in eight 
years},
to \$ 8.50 an hour from \$ 8. 

\EITEM  
Mr. Dinkins also has failed to allay Jewish voters' fears about his
association with the Rev. Jesse Jackson, despite {\em the fact that 
few local non-Jewish politicians have been as vocal for Jewish causes 
in the past 20 years as Mr. Dinkins has}. 

\EITEM  
They wonder whether he has {\em the economic know-how to steer the 
city through a possible fiscal crisis}, and they wonder who will be 
advising him.

\EITEM  
 {\em The appetite for oil-service stocks} has been especially strong , 
although some got hit yesterday when Shearson Lehman Hutton cut its 
short-term investment ratings on them.

\EITEM  
After his decisive primary victory over Mayor Edward I. Koch in
September, Mr. Dinkins coasted, until recently, on a quite comfortable
lead over his Republican opponent, Rudolph Giuliani, {\em the former 
crime buster} who has proved a something of a bust as a candidate.

\EITEM  ``{\em The bottom line is that he is a very genuine and decent 
guy}'', says Malcolm Hoenlein, a Jewish community leader.
\ENDENUMA
\end{EXAMPLE}
In addition, we observed a correlation between {\em larger situation} uses of
definite descriptions (discourse-new, and often hearer-old) and certain
syntactic expressions and lexical items. For example, we noticed that a large
number of uses of definite descriptions in the corpus used for this first
experiment referred to temporal entities such as \LINGEX{the year} or
\LINGEX{the month}, or included proper names in place of the head noun or in
premodifier position, as in \LINGEX{the Querecho Plains of New Mexico} and
\LINGEX{the Iran-Iraq war}. Although these definite descriptions would have
been classified by Hawkins as `larger situation' uses, in many cases they
couldn't really be considered hearer-old or unused: what seems to be happening
in these cases is that the writer assumed the reader would use information
about the visual form of words, or perhaps lexical knowledge, to infer that an
object of that name existed in the world.

We evaluated the strength of these correlations by means of a computer
simulation \cite{vieira&poesio:daarc}. The system attempts to classify the
definite descriptions found in texts syntactically annotated according to the
Penn Treebank format. The system classifies a definite description as
unfamiliar using heuristics based on the syntactic and lexical correlations
just observed, i.e., if either (i) it includes an `unexplanatory modifier',
(ii) it occurs in an apposition or a copular construction, or (iii) it is
modified by a relative clause or prepositional phrase. A definite description
is classified as `larger situation' if its head noun is a temporal expression
such as \LINGEX{year} or \LINGEX{month}, or if its head or premodifiers are
head nouns. The implementation revealed that some of the correlations are very
strong: for example, the agreement between the system's classification and the
annotators' on definite descriptions with a nominal complement, such as
\LINGEX{the fact that \ldots} varied between 93\% and 100\% depending on the
annotator; and on average, 70\% of temporal expressions such as \LINGEX{the
year} were interpreted as larger situation by the annotators.

All of this suggests that in using definite descriptions, writers may not make
just assumptions about their readers's knowledge; they may also rely on their
readers' ability to use lexical or syntactic cues to classify a definite
description as discourse-new even when these readers don't know about the
particular object referred to already. This observation is consistent with
Fraurud's hypothesis that interpreting definite descriptions involves two
processes ---deciding whether a definite description related to some entity in
the discourse or not, and searching the antecedent---and that the two processes
are fairly independent. Our findings also suggest that the classification
process may rely on more than just lexical cues, as Fraurud seems to assume
(taking up a suggestion in \cite{loebner:definites}, see below).

\SECTION{Second Experiment}{second_exp_section}

In order to address some of the questions raised by Experiment 1 we set up a
second experiment.  In this second experiment we modified both the
classification scheme and what we asked the annotators to do.

\SUBSECTION{Revisions to the Annotators' Task}{second_classification_section}

One concern we had in designing this second experiment was to understand better
the reasons for the disagreement among annotators observed in the first
experiment.  In particular, we wanted to understand whether the classification
disagreements reflected disagreements about the final semantic
interpretation. Secondly, in this new experiment we structured the task of
deciding on a classification for a definite description around a series of
questions originating a decision tree, rather than giving our subjects an
explicit preference ranking. A third aspect of the first experiment we wanted
to study more carefully was the distribution of definite descriptions, in
particular, the characteristics of the large number of definite descriptions in
the {\LARGERUNFAMILIAR} class. Finally, we chose truly naive subjects to
perform the classification task.

In order to get a better idea of the extent of agreement among annotators about
the semantic interpretation of definite descriptions, we asked our subjects to
indicate the antecedent in the text for the definite descriptions they
classified as anaphoric or associative. This would also allow us to test how
well subjects did with a `linking' type of classification like the one used in
{\sc muc}-6. We also replaced the {\ANAPHORICSAME} class we had in the first
experiment with a broader {\bf co-referent} class including all cases in which
a definite description is co-referential with its antecedent, whether or not
the head noun was the same: e.g., we asked the subjects to classify as {\bf
co-referent} a definite like \LINGEX{the house} referring back to an antecedent
introduced as \LINGEX{a Victorian home}, which would not have counted as
{\ANAPHORICSAME} in our first experiment.  This resulted in a taxonomy which
was at the same time more semantically oriented and closer to Hawkins' and
Prince's classification schemes: our broadened {\COREFERENT} class coincides
with Hawkins' `anaphoric' and Prince's `textually evoked' classes, whereas the
resulting, narrower `associative' class (that we called {\bf bridging
references}) coincides with Hawkins' `associative anaphoric' and Prince's class
of inferrables. Our intention was to see whether the distinctions proposed by
Hawkins and Prince would result in a better agreement among annotators than the
taxonomy used in our first experiment, i.e., whether the subjects would be more
in agreement about the semantic relation between a definite description and its
antecedent than they were about the relation between the head noun of the
definite description and the head noun of its antecedent.

The {\LARGERUNFAMILIAR} class we had in the first experiment was split back in
two classes, as in Hawkins' and Prince's schemes.  We did this to see whether
indeed these two classes were difficult to distinguish; we also wanted to get a
clearer idea of the relative importance of the two kinds of definites that we
had grouped together in the first annotation. The two classes were called {\bf
larger-situation} and {\bf unfamiliar}.

\subsection{Experimental Conditions}

We used three subjects for Experiment 2. Our subjects were English native
speakers, graduate students of Mathematics, Geography and Mechanical
Engineering at the University of Edinburgh; we will refer to them as C,D, and E
below. They were asked to annotate 14 randomly selected Wall Street Journal
articles, all but one of them different from those used in Experiment 1, and
containing 464 definite descriptions in total.\footnote{The texts are w0766,
wsj\_0003, wsj\_0013, wsj\_0015, wsj\_0018, wsj\_0020, wsj\_0021, wsj\_0022,
wsj\_0024, wsj\_0026, wsj\_0029, wsj\_0034, wsj\_0037, and wsj\_0039.}

Unlike in our first experiment, we did not suggest any relation between the
classes and the syntactic form of the definite descriptions in the
instructions. The subjects were asked to indicate whether the entity referred
to by a definite description i) had been mentioned previously in the text, else
if ii) it was new but related to an entity already mentioned in the text, else
iii) it was new but presumably known to the average reader, or finally iv) it
was new in the text and presumably new to the average reader.

When the description was indicated as discourse-old ({\it i}) or related to
some other entity ({\it ii}), the subjects were asked to locate the previous
mention of the related entity in the text.  Unlike the first experiment, the
subjects did not have the option to classify a definite description as `Idiom';
we instructed them to make a choice and write down their doubts. The written
instructions and the script given to the subjects can be found in Appendix
B. As in Experiment 1, the subjects were given one text to practice before
starting with the analysis of the corpus. They took in average 8 hours to
complete the task.

\subsection{Results}

The distribution of definite descriptions in the four classes according to the
three coders is shown in Table \ref{anns-table2}. We counted all the cases of
doubt separately.

\begin{table}
\begin{tabular}{|l|l|l|l|l|l|l|}
                          & \multicolumn{2}{c|}{C} &
                            \multicolumn{2}{c|}{D} &
                            \multicolumn{2}{c|}{E} \\[1ex] \cline{2-7}
{\bf Class}               & Total & \%    & Total & \%  & Total & \% \\
\hline
{\bf I. Co-referential}   &  205  &  44\% &  211 & 45\% & 201   & 43\%\\
\hline
{\bf II. Bridging}        &  40   & 8.5\% &  29  & 6\%  & 49    & 11\% \\
\hline
{\bf III. Larger situation}&  119 &25.5\% &  115 & 25\% & 93    & 20\% \\
\hline
{\bf IV. Unfamiliar}      &  92   &  20\% &  82  & 18\% & 121   & 26\%\\
\hline
{\bf V. Doubt}           &    8  &  2\%  &  27 &   6\% & 0     & 0\% \\
\hline
{\bf Total}               &  464  &  100\%&  464 & 100\%& 464 & 100\% \\
\hline
\end{tabular}
\caption{\label{anns-table2} Coders' classification of definite   descriptions
in Experiment 2.}
\end{table}

We had 283 cases of complete agreement among annotators on the classification
(61\%): 164 cases of complete agreement on co-referential definite
descriptions, 7 cases of complete agreement on bridging, 65 cases of complete
agreement on larger situation, and 47 cases of complete agreement on the
unfamiliar class. 

As in Experiment 1, we measured the K coefficient of agreement among
annotators; the result for annotators C, D and E is K=0.58 if we consider the
definite descriptions marked as `doubts' (in which case we have 464
descriptions and five classes), K=0.63 if we leave them out (430 descriptions
and the four classes I-IV).

We also measured the extent of agreement among subjects on the antecedents for
co-referential and bridging definite descriptions.  164 descriptions were
classified as {\bf co-referential} by all three coders; of these, 155 (95\%)
were taken by all coders to refer to the same entity (although not necessarily
to the same mention of that entity). 


There were only 7 definite descriptions classified by all three annotators as
{\bf bridging reference}; in 5 of these cases (71\%) the three annotators also
agreed on a textual antecedent (i.e., on the discourse entity to which the
bridging reference was related to).

\subsection{Discussion}

\subsection*{Distribution into classes} 

As shown in Table \ref{anns-table2}, the distribution of definite descriptions
among discourse-new, on the one side, and co-referential with bridging
references, one the other, is roughly the same in Experiment 2 as in Experiment
1, and roughly the same among annotators.  The average percentage of
discourse-new descriptions (larger situation and unfamiliar together) is 46\%,
against an average of 50\% in the first experiment. Having split the
discourse-new class in two in this experiment, we got an indication of the
relative importance of the hearer-old and hearer-new subclasses---about half of
the discourse-new uses fall in each of these classes---but only very
approximate, since the first two annotators classified the majority of these as
{\bf larger-situation}, whereas the last annotator classified the majority as
{\bf unfamiliar}.

As expected, the broader definition of the {\bf co-referent} class resulted in
a larger percentage of definite descriptions being included in this class (an
average of 45\%), and a smaller percentage being included in the {\bf bridging
reference} class. Considering the difference between the relative importance of
the same-head anaphora class in the first experiment and of the co-referent
class in the second experiment we can estimate that approximately 15\% of
definite descriptions are co-referential and have a different head from their
antecedents.

\subsection*{Agreement among annotators}

The agreement among annotators in Experiment 2 was not very high: 61\% total
agreement, which gives K=0.58 or K=0.63, depending on whether we consider
doubts as a class.\footnote{It is difficult to decide what is the best way to
treat cases marked as `doubts'--whether to take them out or to include them as
a separate class--so we give both figures below.} This is worse than the one we
obtained in Experiment 1 (K=0.68 or K=0.73); in fact, this value of K goes
below the level at which we can tentatively assume agreement among the
annotators.

There could be several reasons for the fact that agreeement got worse in this
second experiment. Perhaps the simplest explanation is that we were just using
more classes. In order to check whether this latter was the case, we `merged
back' the classes {\bf larger situation} and {\bf unfamiliar} into one, as we
had in the Experiment 1: that is, we recomputed K after counting all definite
descriptions classified as either {\bf larger situation} or {\bf unfamiliar} as
members of the same class. And indeed, the agreement figures went up from
K=0.63 to K=0.68 (ignoring doubts) when we did so, i.e., back within the
`tentative' margins of agreement according to \cite{carletta:kappa} ($0.68 \leq
x < 0.8$).

The remaining difference between the level of agreement obtained in this
experiment and that obtained in the first one (K=0.73, ignoring doubts) might
have to do with the annotators, with the difficulty of the texts, or with using
a `syntactic' (same head) as opposed to a `semantic' notion of what counts as
co-referential; we are inclined to think that the last two explanations are
more likely. For one thing, we found very few examples of true `mistakes' in
the annotation, as discussed below. Secondly, we observed that the coefficient
of agreement changes dramatically from text to text: in this second experiment,
it varies from K=0.42 to K=0.92 depending on the text, and if we do not count
the worse 3 texts in the second experiment, we get again K=0.73. Third, going
from a `syntactic' to a `semantic' definition of anaphoric definite description
resulted in worse agreement both for co-referential and for bridging
references: looking at the per-class figures, we notice that we went from a
per-class agreement on anaphoric definite descriptions in Experiment 1 of 88\%
to a per-class agreement on coreferential definites of 86\% in Experiment 2;
and the per-class agreement for associative definite descriptions of 59\% went
down rather dramatically to a per-class agreement of 31\% on bridging
descriptions.

The good result obtained by reducing the number of classes led us to try to
find a way of grouping definite descriptions into classes that would result in
a better agreement. An obvious idea was too try with still fewer classes, i.e.,
just two. We first tried the binary division suggested by Fraurud: all {\bf
co-referential} definite descriptions on one side (`subsequent mention'), and
all other definite descriptions on the other (`first mention'). Splitting
things this way did result in an agreement of K=0.76, i.e., within the
`tentative' margins of agreement, although not quite as strong an agreement as
we would have expected. The alternative of putting in one class all
`discourse-related' definite descriptions---{\bf co-referential} and {\bf
bridging} references---and putting {\bf larger situation} and {\bf unfamiliar}
definite descriptions in a second class resulted in a worse agreement, although
by not much (K=0.73).

This suggests that our subjects did reasonably well at distinguishing
first-mention from subsequent-mention entities, but not at drawing more complex
distinctions. They were particularly bad at distinguishing bridging references
from other definite descriptions: dividing the classifications into {\bf
bridging} definites, on the one hand, and all other definite descriptions, on
the other, resulted in a very low agreement (K= 0.24).

We obtained about the same results by computing the `per-class' percentage of
agreement discussed in Section \SECREF{first_exp_section}. The rates of
agreement for each class thus obtained are presented in Table
\ref{table-each-class}. Again, we find that the annotators find it easier to
agree on co-referential definite descriptions, harder to agree on bridging
references; the percentage agreement on the classes {\bf larger situation} and
{\bf unfamiliar} taken individually is much lower than the agreement on the
class {\LARGERUNFAMILIAR} taken as a whole.

\begin{table}
{\small 
\begin{tabular}{|l|l|l|l|l|l|}
{\bf Class} & Total&  Comparisons & Agree & Disagree &\% Agreement \\
\hline
{\bf I. Co-referential}    &  617 & 1234& 1066& 168 & 86\% \\
\hline
{\bf II. Bridging}         &  118 & 236 & 74  & 162 & 31\% \\
\hline
{\bf III. Larger situation}&  327 & 654 & 466 & 188 & 71\% \\
\hline
{\bf IV. Unfamiliar}       &  295 & 590 & 380 & 210 & 64\%\\
\hline
{\bf     Doubt  }          &   35 & 70  & 2   & 68   & 3\% \\
\hline
\end{tabular}
\caption{\label{table-each-class} Per-class agreement in  Experiment 2.}
}
\end{table}
The results in  Table \ref{table-each-class} confirm the indications obtained
by computing agreement for a smaller number of classes: 
our subjects agree pretty much on
{\bf co-referential} definite descriptions, but {\bf bridging references} are
not a natural class.  We discuss the cases of disagreement in more detail next.

\subsection*{Classification  disagreements}

There are two basic kinds of disagreements among annotators: about
classification, and about the identification of an antecedent. 

There were 29 cases of complete classification disagreement among annotators,
i.e., cases in which no two annotators classified a definite description in the
same way, and 144 cases of partial disagreement. All four of the possible
combinations of total disagreement were observed, but the two most common
combinations were BCU (bridging, co-referential, and unfamiliar) and BLU
(bridging, larger situation, and unfamiliar); all six combinations of partial
disagreements were also observed.  As we do not have the space for discussing
each case in detail, we will concentrate on pointing out what we take to be the
most interesting observations, especially from the perspective of designing a
corpus annotation scheme for anaphoric expressions.

We found very few true `mistakes'. We had some problems due to the presence of
idioms such as \LINGEX{they had to pick up the slack} or \LINGEX{on the whole
the situation was better than expected}. But in general, most of the
disagreements were due to genuine problems in assigning a unique classification
to definite descriptions.

The `mistakes' that our annotators did make were of the form exemplified by
\SREF{class:mistake}. In this case, all three annotators indicate the same
antecedent (\LINGEX{the potential payoff}) for the definite description
\LINGEX{the rewards}, but whereas two of them classify \LINGEX{the rewards} as
{\bf co-referential}, one of them classifies it as {\bf bridging}. What seems
to be happening here and in similar cases is that even though we asked the
subjects to classify `semantically,' they ended up using a notion of
`relatedness' which is more like the notion of `associative' in Experiment
1. (We found 10 such cases of partial disagreement between bridging and
co-referential in which all three subjects indicated the same antecedent for
the definite description.)

\begin{EXAMPLE}
\ENEW{class:mistake} New England Electric System bowed out of the bidding for
Public Service Co. of New Hampshire, saying that the risks were too high and
\LINGEX{the potential payoff} too far in the future to justify a higher offer.

\ldots

``When we evaluated raising our bid, the risks seemed substantial and
persistent over the next five years, and \LINGEX{the rewards}
seemed a long way out.''
\end{EXAMPLE}
A particularly interesting version of this problem appears in the following
example, when two annotators took the verb \LINGEX{to refund} as antecedent of
the definite description \LINGEX{the refund}, but one of them interpreted the
definite as co-referential with the eventuality, the other as bridging.

\begin{EXAMPLE}
\ENEW{class:mistake:2}
Commonwealth Edison Co. was ordered \LINGEX{to refund} about \$250 million to
its current and former ratepayers for illegal rates collected for cost overruns
on a nuclear power plant. 

\LINGEX{The refund} was about \$55 million more than previously ordered by the
Illinois Commerce Commission and trade groups said it may be the largest ever
required of a state or local utility.
\end{EXAMPLE}

As could be expected by the discussion of the K results above, the most common
disagreements (35 cases of partial disagreement out of 144) were between the
classes {\bf larger situation} and {\bf unfamiliar}. One typical source of
disagreement was the `introductory' use of definite descriptions, common in
newspapers: thus, for example, some of our annotators would classify
\LINGEX{the Illinois Commerce Commission} as larger situation, other as
unfamiliar. In many cases in which this form of ambiguity was encountered, the
definite description worked effectively as a proper name: \LINGEX{the
world-wide supercomputer law}, \LINGEX{the new US trade law}, or \LINGEX{the
face of personal computing}.

Rather surprisingly, from a semantic perspective, the second most common form
of disagreement was between the {\bf co-referential} and {\bf bridging}
classes. In this case, the problem typically was that different subjects would
choose different antecedents for a certain definite description. Thus, in
example \SREF{class:mistake:2}, the third annotator indicated \LINGEX{\$250
million} as the antecedent for \LINGEX{the refund}, and classified the definite
description as co-referential. A similar example is \SREF{class:mistake:3}, in
which two of the annotators classified \LINGEX{the spinoff} as bridging on
\LINGEX{spinoff Cray Computer Corp.}, whereas the third classified it as
co-referential with \LINGEX{the pending spinoff}.

\begin{EXAMPLE}
\ENEW{class:mistake:3}
The survival of \LINGEX{spinoff Cray Computer Corp.} as a fledgling in the
supercomputer business appears to depend heavily on the creativity -- and
longevity -- of its chairman and chief designer, Seymour Cray.  

\ldots

Documents filed with the Securities and Exchange Commission on \LINGEX{the
pending spinoff}  disclosed that Cray Research Inc. will withdraw the almost
\$100 million in financing it is providing the new firm if Mr. Cray leaves or
if the product-design project he heads is scrapped. 

\ldots

While many of the risks were anticipated when Minneapolis-based Cray Research
first announced \LINGEX{the spinoff} in May, the strings it attached to the
financing hadn't been made public until yesterday. 
\end{EXAMPLE} 

An example of total (BLU) disagreement is the following: 

\begin{EXAMPLE}
\ENEW{blu:ex} 
Mr. Rapanelli recently has said \LINGEX{the government of President Carlos
Menem, who took office July 8,} feels a significant reduction of principal and
interest is the  only way the debt problem may be solved.
\end{EXAMPLE}
In this case, we can see that all three interpretations are acceptable: 
we may take the definite description \LINGEX{the government of President Carlos
Menem, who took office July 8,} either as a case of bridging reference on the
previously mentioned \LINGEX{Argentina}, or as a larger situation use, or as a
case of unfamiliar definite description, especially if we assume that this
latter class coincides with Prince's containing inferrables.

In conclusion, our figures can be seen as an empirical verification of
Fraurud's and Prince's hypothesis that the classification disagreements among
annotators depend to a large extent on the task they are asked to do, rather
than reflecting true differences in semantic intuitions.

\subsection*{Antecedent disagreements}

Interestingly, we also found cases of disagreement about the antecedent of a
definite description.

We have already discussed the most common case of antecedent disagreement: this
is the case in which a definite description could equally well be taken as
co-referential with one discourse entity or as bridging to another: for
example, in an article in which the writer starts discussing {\em Aetna Life \&
Casualty}, and then goes on mentioning {\em major insurers}, either discourse
entity could then serve as `antecedent' for the subsequent definite description
\LINGEX{the insurer}, depending on whether the definite description is
classified as co-referential or bridging.

Perhaps most interesting of all cases of disagreement about the antecedent are
examples such as \SREF{undersp:bridging}. One subject indicated \LINGEX{parts
of the factory} as the antecedent; another indicated \LINGEX{the factory}; and
the third indicated \LINGEX{areas of the factory}.

\begin{EXAMPLE}
\ENEW{undersp:bridging}
About 160 workers at \LINGEX{a factory} that made paper for the Kent filters
were exposed to asbestos in the 1950s.  \LINGEX{Areas of the factory} were
particularly dusty where the crocidolite was used.  Workers dumped large burlap
sacks of the imported material into a huge bin, poured in cotton and acetate
fibers and mechanically mixed the dry fibers in a process used to make
filters. Workers described "clouds of blue dust" that hung over \LINGEX{parts
of the factory}, even though exhaust fans ventilated \LINGEX{the area}.

\end{EXAMPLE}
What's  interesting about this example is that the text does not provide us
with enough information to decide about the correct interpretation; it is as if
the writer didn't think it necessary for the reader to assign an unambiguous
interpretation to the definite description. Similar cases of `underspecified'
definite descriptions have been observed before (e.g., Nunberg's \LINGEX{John
shot himself in the foot} or \LINGEX{I'm going to the store} mentioned in
\cite{clark&marshall:81}) but no real account has been given of the conditions
under which they are possible.

\SECTION{Discussion and Conclusions}{conclusions_section}

\subsection{Some Consequences}

\subsection*{Consequences for Corpus Annotation}

This study raises the issue of how feasible it is to annotate corpora for
anaphoric information. We observed two problems about the task of classifying
definite descriptions: first, neither of the more complex classification
schemes we tested resulted in a very good agreement among annotators; and
second, even the task of identifying the antecedent of `discourse-related'
definite descriptions (i.e., co-referential and bridging) is problematic---we
only obtained an acceptable agreement in the case of co-referential definite
descriptions, and it was difficult for our annotators to choose a single
antecedent for a definite description when both bridging and co-reference are
allowed. These results indicate that annotating corpora for anaphoric
information may be more difficult than expected. The task of indicating a
unique antecedent for bridging definite descriptions appears to be especially
challenging, for the reasons discussed above (multiple equally good antecedents
and referential underspecification, for example).

On the positive side, we have two positive observations: subjects do reasonably
well at distinguishing first-mention from subsequent-mention antecedents, and
at identifying the antecedent of a subsequent-mention definite description.  A
classification scheme based on this distinction (such as Fraurud's) and that
just asked subjects to indicate an antecedent for subsequent-mention definite
descriptions may have a chance of resulting in a standardized annotation.  Even
in this case, however, the agreement we observed was not very high.

The possibility we are exploring is that these results might get better if
annotators are given computer support in the form of a semi-automatic
classifier--i.e., a system capable of suggesting to annotators a classification
for definite descriptions, including possibly an indication of how reliable the
classification might be. We briefly discuss below our progress in this
direction so far.

\subsection*{Consequences for Linguistic Theory}

Our study confirms the findings of previous work (e.g., \cite{fraurud:90}) that
a great number of the definite descriptions in texts are discourse-new: in our
second experiment we found an equal number of discourse-new and
`discourse-related' definite descriptions, although many of the definite
descriptions classified as discourse new could be seen as associative in a
loose sense. Interestingly, this suggests that each of the competing hypotheses
about the licensing conditions for definite descriptions-- the uniqueness and
the familiarity theory-- accounts satisfactorily for about half of the data.

Of the existing theories of definite descriptions, the one that comes closest
to accounting for all of the uses of definite descriptions that we observed is
L{\"{o}}bner's \shortcite{loebner:definites}. L{\"{o}}bner proposes that the
defining property of definite descriptions, from a semantic point of view, is
that they indicate that the head noun complex denotes a \NEWTERM{functional
concept}, i.e., a function (which, according to L{\"{o}}bner, can take one, two
or three arguments). He argues that some head noun complexes denote such a
function on purely lexical semantic grounds: this is the case, for example, of
the head noun complexes in \LINGEX{the father of Mr. Smith}, \LINGEX{the first
man to sail to America} and \LINGEX{the fact that life started on Earth}; he
calls these definite descriptions \NEWTERM{semantic definites}. In other cases,
such as \LINGEX{the dog}, the head noun by itself would not denote a function,
but a sort: in these cases, according to L{\"{o}}bner, the use of a definite
dscription is only felicitous if context indicates the function to be
used. This latter class of \NEWTERM{pragmatic definites} includes the
best-known cases of familiar definites--anaphoric, immediate and visible
situation, and larger situation--as well as some cases classified by Hawkins as
unfamiliar and by Prince as containing inferrables. L{\"{o}}bner does not
discuss the conditions under which a writer can assume that the reader can
recognize that context creates a functional concept out of a sortal one, but
his account could be supplemented by Clark and Marshall's theory of what may
count as a basis for a mutual knowledge induction schema
\cite{clark&marshall:81}.\footnote{L{\"{o}}bner's theory still does not account
for generic uses of definite descriptions.}

\subsection*{Consequences for Processing Theories}

Given that first-mention definite descriptions are so numerous, and that
recognizing them does not depend on commonsense knowledge alone, we conclude
that any general theory of definite description interpretation should include
methods for recognizing such definites. The architecture of our own classifier
(see below) is also consistent with Fraurud's hypothesis that these methods are
not just used when no suitable antecedent can be found, but more extensive
investigations will be needed before we can conclude that this architecture
significantly outperforms other ones.

The presence of such a large number of discourse-new definite descriptions is
also problematic for the idea that definite descriptions are interpreted with
respect to the global focus \cite{grosz:thesis,grosz&sidner:86}. A significant
percentage of the larger situation definite descriptions encountered in our
corpus cannot be said to be in the `global focus' in any significant sense: as
we observed above, in many of these cases the writer seems to rely on the
reader's capability to add a new object such as \LINGEX{the Illinois Commerce
Commission} to her/his model of the world, rather than expecting that object to
be already present.

\subsection{A (Semi)-Automatic Classifier}

As already mentioned, we are in the course of implementing a system capable of
performing the classification task. The idea is to help the human classifiers
in their task by suggesting possible classifications, and possible antecedents
in the case of discourse-related definite descriptions.

Our system implements the `dual-processing' strategy discussed above.  On the
one hand, it attempts to resolve anaphoric same-head definite descriptions by
maintaining a simple discourse model and searching back into this model to find
all possible antecedents of a definite description (using a special matching
heuristic to deal with pre- and post-modification). On the other, it uses
heuristics to identify unfamiliar and larger situation definite descriptions on
the basis of syntactic information and very little lexical information about
nouns that take complements. The current order of application of the resolution
and classification steps has been determined by empirical testing, and has been
compared with that suggested by decision-tree learning techniques.

We `trained' a version of the system on the corpus used for the first
experiment, and then compared its classification of the corpus used for the
second experiment with that of our three subjects.\footnote{As the two
classification schemes were different, the comparison involved a conversion of
the annotations produced in the second experiment into ones using the scheme
used in the first experiment.}  We developed two versions of the system: one
which only attempts to classify subsequent mention and discourse-new definite
descriptions \cite{vieira&poesio:daarc}, and one which also attempts to
classify bridging references \cite{poesio-et-al:ACL97}. 

The first version of the system finds a classification for 318 definite
descriptions out of the 464 in our test data (the articles used in the second
experiment). The agreement between the system and the three annotators on the
two classes first mention and subsequent mention is K=0.70 overall (K=0.77 for
the three annotators on the converted annotation), if all definite descriptions
to which the system can't assign a classification are treated as first-mention;
the coefficient of agreement is K=0.78 if we do not count the definite
descriptions that the system cannot classify (K=0.81 for the annotators on just
those definite descriptions).

The version of the system that also attempts to recognize bridging references
has a worse performance, which is not surprising given the problems our
subjects had in classifying bridging descriptions. This version of the system
finds a classification for 355 descriptions out of 464, and its agreement with
the three annotators is K=0.63 if the cases that the system cannot classify are
not counted (K=0.70 for the three annotators on 3 categories with just these
definites); K=0.57 if we count the cases that the system does not classify as
discourse-new (for 447 descriptions); and K=0.63 again if we count the cases
that the system does not classify as bridging (again, 447 descriptions).

\subsection{Future Work}

We collected plenty of data about definite descriptions that we are still in
the process of analyzing. One issue we are studying at the moment is what to do
with bridging references: how to classify them if at all, and how to process
them.  We also intend to study Loebner's hypothesis about the role played by
the distinction between `sortal' and `relational' head nouns in determining the
type of process involved in the resolution of a definite description, possibly
by finding a way to ask our subjects to recognize these distinctions. And we
plan to study the issue of generic definites.

An obvious direction in which to extend this study is by looking at other kinds
of anaphoric expressions such as pronouns and demonstratives. We are performing
preliminary studies in this direction.

Finally, we would like to emphasize that although this study is the most
extensive investigation of definite description use in a corpus that we know of
(we looked at a total of more than 1400 definite descriptions in 33 texts,
i.e., almost three times as many as in Fraurud's study), in practice we still
got very little data on many of the uses of definite descriptions, so some
caution is necessary in interpreting these results. The problem is that the
kind of analysis we performed is extremely time consuming: it will be crucial
in the future to find ways of performing this task that will allow us to
analyze more data, possibly with the help of computer simulations.

\section*{Acknowledgments}

We wish to thank Jean Carletta for much help both with designing the
experiments and with the analysis of the results. We are also grateful to Ellen
Bard, Robin Cooper, Kari Fraurud, Janet Hitzeman, Kjetil Strand, and our
anonymous reviewers for many helpful comments. Massimo Poesio holds an Advanced
Research Fellowship from EPSRC.


\bibliographystyle{fullname}

\newpage

\appendix

\section{Instructions to the Annotators (First Experiment)}

\begin{center}
{\bf Classification of uses of ``the''-phrases}
\end{center}

You will receive a set of texts to read and annotate. From the texts, the
system will extract and present you ``the''-phrases and will ask you for a
classification.  You must choose one of the following classes: \vspace{1mm}

{\bf 1. ANAPHORIC} (same noun):
 For anaphoric ``the''-phrases  the text presents an antecedent
noun phrase 
which has  the same noun of the given ``the''-phrase. The
interpretation of the given  ``the''-phrase is based on this  previous
noun-phrase. 
\vspace{1mm}

{\bf 2. ASSOCIATIVE:}
For associative ``the''-phrases  the text presents an antecedent
noun phrase which has a different noun for the interpretation of the
given ``the''-phrase. The antecedent for the ``the''-phrase in this
case  may 

a) allow  an inference towards  the interpretation of the
``the''-phrase,

b) be a synonym,

c) be  an associate such as part-of,  is-a, etc. 

d) a proper name

\vspace{1mm}
{\bf 3. LARGER SITUATION/UNFAMILIAR:}
 For larger situation   use of ``the''-phrases  {\bf you do not find an
explicit antecedent in the text}, because the reference is based on 
basic common knowledge: 

a)  first occurrences  of proper names (subsequent occurrences must be
considered as anaphoric),

b)  reference to times,

c) community common knowledge;

d) proper names in premodifier position.

Also for unfamiliar uses of ``the''-phrases  {\bf the text does not  
provide an antecedent}. The ``the''-phrase refers to
something {\bf  new} to the text.  The help for the interpration may
be given together with the ``the''-phrase as in

e) restrictive relative clauses (the ... that ... - RC in general)

f) associative clauses (the ...  of ... - PP in general)

g) NP complements (the fact that ...,the conclusion that ...)

h) unexplanatory modifiers (the first ..., the best ...)

i) appositive structures (James Dean , the actor)

j) copulas (the actor is James Dean) 

\vspace{1mm}
{\bf 4. IDIOM:}
 ``The''-phrases can be used just as idiomatic expressions, indirect
references or metaphorical uses.

\vspace{1mm}
{\bf 5. DOUBT:} 
 When you are in doubt about the classification:
  a comment on your doubt is requested.

 \textsc{Preference order for the classification:}
In spite of the fact that definites  often  fall in  more than one class 
of use,   the identification of a unique class is required. In
order to make the choices uniform, priority is to be given to anaphoric
situations.
 According to this ordering, cases like  ``the White House'' or
``the
government''  are
anaphoric rather than larger situation,
{\bf when it has already occurred once in the text}. When a
``the''-phrase seems to belong both to larger sit./unfamiliar and
associative classes, preference is given to larger sit./unfamiliar.

{\bf Examples}

[Examples from the corpus were given as  in section 
\ref{first_exp_section}.]

\newpage

\begin{center}
{\bf Summary}
\end{center}

\
\begin{tabular}{ll}
WHEN AN ANTECEDENT IS~~~~~          &WHEN THE REFERENT FOR \\
GIVEN EXPLICITLY IN THE             &THE DESCRIPTION IS\\
\vspace{2mm}
TEXT:(1,2)                          &KNOWN OR NEW:(3,4)\\
1.: ANAPHORIC                       &3.: LARGER SIT./UNFAMILIAR \\
There is an antecedent in the       &The ``the''-phrase is novel in\\
text which has the same             &the text, unique identifiable,\\
descriptive noun of the             &or based on common knowlege\\
``the''-phrase.                     
\vspace{2mm} 
                                    & or is given with its referent\\
2.: ASSOCIATIVE                     &4.: IDIOM\\
There is an antecedent in the       &The ``the''-phrase is an\\
text which has a different noun,    &idiomatic expression \\
but it is a synonym or  associate   & \\
\vspace{2mm}
to the description.                 &\\
\end{tabular}

\begin{enumerate}
\item
  \begin{enumerate}
  \item a house: {\bf the house}
  \end{enumerate}
\item
   \begin{enumerate}
   \item something has changed: {\bf the change}
   \item a home: {\bf the house}
   \item a house: {\bf the door}
   \item Kadane Co.: {\bf the company}  
   \end{enumerate}
\item
   \begin{enumerate}
   \item the White House (first occurrence)
   \item the third quarter  
   \item the nation
   \item the Iran-Iraq war
   \item {\bf the woman} he likes
   \item {\bf the door} of the house 
   \item {\bf the fact} that
   \item the first, the best, the highest, the tallest ... 
   \item James Dean, {\bf the actor}
   \item {\bf the actor} is James Dean
   \end{enumerate}
\item
   \begin{enumerate}
   \item back into {\bf the soup}
   \end{enumerate}
\end{enumerate}

\newpage

\SECTION{Instructions to the Subjects (Second Experiment)}{appendix_B_section}

\begin{center} {\bf Text Annotation of Definite Descriptions} 
\end{center}

This material provides you with instructions, examples and some training for
the text-annotation task.  The task consists of reading newspaper articles and
analysing occurrences of \textsc{definite descriptions}, which are expressions
starting with the definite article \textsc{the}.  We will call these
expressions DDs or DD. DDs describe things, ideas or entities which are talked
about in the text. The things, ideas or entities being described by DDs will be
called \textsc{entities}. You should look at the text, carefully in order to
indicate whether the \textsc{entity} was mentioned before in the text and if
so, to indicate where.  You will receive a set of texts and their corresponding
tables to fill in.  There are basically four cases to be considered:
\vspace{5mm}

{\bf 1.} Usually DDs pick up an entity introduced before in the
text. For instance, in the sequence:

``{\em Mrs. Park}
is saving  to buy  an apartment. {\em The housewife} is saving harder
than ever.''

the \textsc{entity} described by the DD ``{\em the
  housewife}''  was mentioned  before as ``{\em Mrs. Park}''. \\

{\bf 2.} If the \textsc{entity} itself was not mentioned before but its
 interpretation is based on , dependent on, or related to some other idea or
 thing in the text, you should indicate it. For instance, in the sequence:

`` The Parks wanted to buy {\em an apartment} but {\em the price} was very
high.

 the \textsc{entity} described by the DD {\em the price} is
related to the idea expressed by {\em an apartment} in the text. \\

{\bf 3.}  It may also be the case that the DD was not mentioned before and is
not related to something in the text, but it refers to something which is part
of the common knowledge of the writer and readers in general. (The texts to be
analysed are Wall Street Journal articles - location and time, for instance,
are usually known to the general reader from sources which are outside the
text). Example:

 ``During {\em the past 15 years}  housing prices increased nearly fivefold''.

here, the \textsc{entity} described by the DD {\em the past 15 years} is known to the general reader of the Wall Street Journal and
  was not mentioned before in the text. \\

{\bf 4.}  Or it may be the case that the DD is self-explanatory or it is given
together with its own identification. In these cases it becomes clear to the
general reader what is being talked about even without previous mention in the
text or without previous common knowledge of it. For instance:

``The proposed legislation is aimed at rectifying some of {\em the inequities in the current land-ownership system}.''

the \textsc{entity} described here is new in the text, and is not part
of the knowledge of readers but the DD {\em the inequities in the
  current land-ownership system} is self-explanatory. \\

\vspace{3mm}

The texts will be presented to you in the following format: on the
left, the text with its DDs in evidence; on the right, the keys
(number of the sentence/number of DD)  and the  DD 
to be analysed. The key is for internal control only, but it may help you
to find DDs in the table you have to fill in.\\
\footnotesize

\noindent {\bf Text 0}

\noindent 
1 Y. J. Park and her family scrimped for four   
~~~~~~~~{\bf  (1/1) the price} \\
years to buy a tiny apartment here, but found \\
that   the closer they got to saving  the \$40,000 \\
 they originally needed,   the more  {\bf the price}   \\
rose. \\
...\\
\noindent 
3 Now  {\bf the 33-year-old housewife}, whose  
~~~~~~~~~~ {\bf (3/2) the 33-year-old housewife} \\
husband earns a modest salary as an assistant \\
 professor of economics, is saving harder \\
than ever. \\
...

\noindent 
9 During {\bf the past 15 years}, the report showed, 
~~~~{\bf (9/3) the past 15 years}\\
housing prices increased nearly fivefold. \\
...

\noindent 
22 The proposed legislation is aimed at rectifying 
~~~~~{\bf (22/4) the inequities in the current} \\
some of {\bf the inequities in the current land-}
~~~~~~~{\bf land-ownership system}\\
{\bf ownership system}.

\normalsize
\hspace{2mm}

 You can draw arrows, use colours, whatever you like over
the text and the list of DDs to help   your
analysis and then you should complete a  table in the format below.

\begin{table}[h]
\begin{tabular}{|l|l|l|l|l|}
\hline
{\bf  Text 0} & {\bf \textsc{Definite description}} & {\footnotesize \bf
 LINK} &  {\bf \footnotesize LINK} &
 {\footnotesize \bf  NO}\\
  & &   & {\bf Sentence no./} & {\footnotesize \bf LINK}\\
 
 {\bf Key} & & {\footnotesize \bf =/R}&{\bf previous mention}& {\footnotesize \bf K/D}\\
\hline
{\bf \footnotesize 1/1} & {\footnotesize \bf the price} &  &  &  \\
\hline

{\bf \footnotesize  3/2} & {\footnotesize \bf the 33-year-old housewife} &  & &   \\
\hline 
\vdots & & & & \\
\hline
\end{tabular}
\end{table}

Each case (1 to 4, above) is to be indicated on the table according to the following
(see examples in the table below):

Whenever you find a previous mention in the text of the DD you should mark the
column {\bf LINK}:\\

{\bf 1.} Mark  ``=''  if the \textsc{entity} described was mentioned before.\\

{\bf 2.} Mark ``R'' if the \textsc{entity} described is new but it is related/based/dependent on
something  mentioned before). \\

In the case of both {\bf 1} and {\bf 2} you should provide the
sentence number where the previous/related mention is  and write down the
previous/related mention of it (see example in the table below).\\

 If the entity was not previously mentioned in the text and it is
  not related to something mentioned before, then mark the column {\bf NO LINK}: \\

{\bf 3.} Mark ``K'' if  it is  something of writer/readers'
common knowledge. \\

{\bf 4.} Mark ``D'' if  it is  new in the
text and the readers have no previous knowledge about it but the
description is enough to make readers identify it.

\begin{table}[h]
\begin{tabular}{|l|l|l|l|l|}
\hline
{\bf  Text 0} & {\bf \textsc{Definite description}} & {\footnotesize \bf
 LINK} &  {\bf \footnotesize LINK} &
 {\footnotesize \bf  NO}\\
  & &   & {\bf Sentence no./} & {\footnotesize \bf LINK}\\
{\bf  Key} & & {\footnotesize \bf =/R}& {\bf previous mention} & {\footnotesize \bf K/D}\\
\hline
{\bf \footnotesize 1/1} & {\footnotesize \bf  the price} & R & {\footnotesize \em 1/apartment} &  \\
\hline

{\bf \footnotesize 3/2} & {\footnotesize \bf  the 33-year-old housewife} & = &
{\footnotesize \em 1/Y.J. Park} &   \\
\hline 
 
{\bf \footnotesize 9/3} & {\footnotesize \bf  the past 15 years} & & &K\\ 

\hline
{\bf \footnotesize 22/4} & {\footnotesize \bf the inequities in the current } & &
---& \\
& {\footnotesize \bf land-ownership system}  & &--- & D\\ 
\hline
\end{tabular}
\end{table}
In case of doubt just leave the line in blank and comment at the back of the
page using the key number to identify the DD you are commenting on.

\begin{center}
{\bf Examples}
\end{center}

Next we present some examples and further explanation for each one of
the four cases that are bein considered.

\noindent {\bf Case 1 - LINK (=)}

For case no. 1 you may find a previous mention that may be equal or
different from the DD ( for instance, \underline{the government} - {\bf
  the government}, \underline{a report} - {\bf the
report}, and  \underline{three bills} - {\bf the proposed legislation} in
the examples below); distances
from previous mentions and DDs may also vary.

\begin{itemize}

\item
 Meanwhile,  \underline{the government}'s Land Bureau reports that only
about a third of Korean families own their own homes.
 Last week,  {\bf the government} took three bills to   the National Assembly.

\item  
Last May, a government panel released \underline{a report} on   the extent and
causes of  the problem. 
During the past 15 years,  {\bf the report} showed, housing prices increased nearly fivefold.
 
\item 
Last week, the government took \underline{three bills} to the National Assembly.
{\bf The proposed legislation} is aimed at rectifying some of 
the inequities in the current land-ownership system.

\end{itemize}

\noindent {\bf Case 2 - LINK (R)}

Here are cases of DDs which are related to something that
was present in the text. If you ask for the examples below, ``Which
{\em government, 
population, nation} is that?'',``Which {\em blame} is that?'' the answer is given by
something previously mentioned in the text (Koreans, and the increase of
housing prices, respectively) \footnote{Note that  DDs like
  {\em the blame, the government, the population}, 
which are  case 2 in their first occurrence,
are to be considered case 1 in possible posterior occurrences.}.

\begin{itemize}

\item For  the Parks and millions of other young \underline{Koreans},  the long-cherished dream of home ownership has become a cruel illusion.
For {\bf the government}, it has become a highly volatile political
issue.
In 1987, a quarter of  {\bf the population} owned 91\% of  {\bf the
  nation}'s 71,895 square kilometers of private land.

\item
 During   the past 15 years,  the report showed, \underline{housing prices increased nearly fivefold}.
  The report laid  {\bf the blame} on speculators, who it said
had pushed land prices up ninefold.

\end{itemize}
 
\noindent {\bf Case 3 - NO LINK (K)}

These cases of DDs are based on the common reader's knowledge. The texts to be
analysed are Wall
Street Journal articles - location and time, for instance, are
usually known to the general reader from sources which are outside the
text \footnote{Note that a DD like ``the government'' may belong to case 2 as
exemplified, but it may refer to the U.S.A. in another text,
without any explicit mention of U.S.A in the text, since it is the
country where the newspaper is produced.  In such a situation the DD
``the government'' belongs to case 3. It may also  be the case
that the  entity  is  part of the readers'
  knowledge but was mentioned before, in this situation it belongs to case 1.}.

\begin{itemize}

\item
 For example , officials at Walnut Creek office learned that   the
Amfac Hotel near {\bf the San Francisco airport}, which is insured by
Aetna, was badly damaged when they saw it on network television news.

\item
Adjusters who had been working on  {\bf the East Coast} say  the
insurer  will still be processing claims from that storm through December .

\end{itemize}

\noindent {\bf Case 4 - NO LINK (D)}

These cases of DDs are self-explanatory or accompained by their
identification. For instance if you ask ``Which {\em difficulty} is
that?'',
``Which {\em fact} is that?'', ``Which {\em know-how} is that?''
etc. for the examples below, the answer is
given by the DD itself. In the last example the DD is accompained by
its explanation.

\begin{itemize}

\item
 Because of {\bf the difficulty of assessing the damages caused by
  the earthquake}, Aetna pulled together a team of its most experienced claims adjusters from around the country .

\item
 They wonder whether he has {\bf the economic know-how to steer the
 city through a possible fiscal crisis}.

\item 
Mr. Dinkins also has failed to allay Jewish voters' fears about his
association with the Rev. Jesse Jackson, despite {\bf the fact that
  few local non-Jewish politicians have been as vocal for Jewish
  causes in the past 20 years as Mr. Dinkins has}.

\item But racial gerrymandering is not {\bf the best way to accomplish that essential goal}.

\item {\bf The first hybrid corn seeds produced using this mechanical approach} were introduced in the 1930s and they yielded as much as 20 \% more corn than naturally pollinated plants.

\item {\bf The Citizens Coalition for Economic Justice},{\em  a public-interest group leading the charge for radical reform}, wants restrictions on landholdings, high taxation of capital gains, and drastic revamping of the value-assessment system on which
 property taxes are based.

\end{itemize}

\noindent {\bf SCRIPT}

\vspace{5mm}

In order to help you filling in the table, answer the YES-NO questions
below
for each one of the DDs in the text. When the answer
for the question is YES (Y) you have an action to follow, if the answer is
NO (N), skip to the next question.

\begin{enumerate}
\item {\normalsize \bf  Does the DD describe an \textsc{entity}   mentioned before?}
\begin{itemize}
\item[{\bf Y}] Mark ``{\bf \small  =}'' (column LINK) to indicate
    that the same entity was mentioned before
    and tell where by providing  the sentence number and the 
    words used in the  previous mention.
\item [{\bf N}] Go to question no. 2.
\end{itemize}
\item {\normalsize \bf Is the \textsc{entity} new but related to something
  mentioned berfore? If you ask: ``Which entity is that?'', is the answer
  based on previous text \footnote{ 
  For instance if you ask: ``Which {\em price} is
  that?'' for {\em the price} in sentence number 1, given above, your answer is
  based on {\em apartment} in the text.}?} \\
\begin{itemize}
\item [{\bf Y}] Mark ``{\bf \small R}'' (column LINK) to indicate related entity
  and provide the sentence number and 
    the  previous mention  on which the DD  is based .
\item [{\bf N}] Go to  question no. 3.
\end{itemize}
\item {\normalsize Is the \textsc{entity} new in the text? If it was not mentioned before and its interpretation is
  not based on the previous text, then: {\bf is it something mutually known by
  writer and general readers of the Wall Street Journal?}}
\begin{itemize}
\item [{\bf Y}] Mark ``{\bf \small K}'' (column NO LINK) to indicate
  general knowledge about the entity.
\item [{\bf N}] Go to question no. 4.
\end{itemize}
\item  {\normalsize Is the \textsc{entity} new in the text? If it was not mentioned before and its interpretation is
  not based on the previous text, then: {\bf is it self-explanatory or
  accompanied by its identification?}}
\begin{itemize}
\item [{\bf Y}] mark ``{\bf \small D}'' (column NO LINK) to indicate that the
  description is enough to make readers identify the entity.
\item [{\bf N}] Leave the line in  blank and comment at the back of the
  page using the key number to identify the DD.''
\end{itemize}
\end{enumerate}

\end{document}